\def\dajpreprint{}

\documentclass[11pt]{article}

\PassOptionsToPackage{table}{xcolor}
\ifdefined\dajpreprint
  \usepackage[preprint]{acl}
\else
  \usepackage[review]{acl}
\fi
\usepackage{times}
\usepackage{latexsym}

\usepackage[utf8]{inputenc}
\usepackage[T1]{fontenc}
\usepackage{microtype}
\usepackage{inconsolata}
\usepackage{graphicx}
\graphicspath{{figures/}}
\usepackage{subcaption}
\usepackage{wrapfig}
\usepackage{tabularx}
\usepackage{booktabs}
\usepackage{multirow}
\usepackage{xcolor}
\usepackage{pifont}
\newcommand{\cmark}{\textcolor{green!48!black}{\ding{51}}}
\newcommand{\xmark}{\textcolor{red!75!black}{\ding{55}}}
\newcommand{\pmark}{\textcolor{orange!85!black}{\ding{51}\kern-1.05ex\raisebox{.55ex}{\rotatebox[origin=c]{125}{\rule{1.25ex}{0.6pt}}}}}
\usepackage{url}
\usepackage{makecell}
\usepackage{listings}
\usepackage{upquote}
\usepackage{amsmath}
\usepackage{amssymb}
\usepackage{mathtools}
\usepackage{amsthm}
\usepackage[capitalize,noabbrev]{cleveref}
\usepackage[most]{tcolorbox}
\tcbuselibrary{listings}
\definecolor{promptaccent}{HTML}{6A9BCC}
\definecolor{prompthead}{HTML}{2F5F8F}
\definecolor{promptslot}{HTML}{B5532E}
\definecolor{prompttext}{HTML}{333333}
\lstdefinestyle{prompt}{
  basicstyle=\ttfamily\fontsize{7.4}{9.4}\selectfont\color{prompttext},
  breaklines=true,
  breakatwhitespace=true,
  postbreak={},
  breakindent=1.5em,
  columns=fullflexible,
  keepspaces=true,
  tabsize=2,
  showspaces=false,
  showstringspaces=false,
  showtabs=false,
  upquote=true,
  aboveskip=0pt,
  belowskip=0pt,
  moredelim=[s][\color{prompthead}\bfseries]{===}{===},
  moredelim=[s][\color{prompthead}\bfseries]{---}{---},
  moredelim=[l][\color{prompthead}\bfseries]{STEP},
  moredelim=[l][\color{prompthead}\bfseries]{EVALUATION},
  moredelim=[l][\color{prompthead}\bfseries]{OUTPUT},
  moredelim=[s][\color{promptslot}]\{\},
  moredelim=[s][\color{promptslot}\bfseries]{<selection>}{</selection>},
}
\newtcolorbox[auto counter, crefname={listing}{listings}, Crefname={Listing}{Listings}]{promptframe}[2][]{%
  enhanced,
  colback=promptaccent!4,
  colframe=promptaccent,
  boxrule=0.6pt,
  arc=4pt,
  attach boxed title to top left={xshift=8pt, yshift=-\tcboxedtitleheight/2},
  boxed title style={colback=prompthead, colframe=prompthead, arc=2pt, boxrule=0pt},
  coltitle=white,
  fonttitle=\bfseries\small,
  title=Listing~\thetcbcounter: #2,
  left=10pt, right=10pt, top=12pt, bottom=6pt,
  #1
}
\usepackage[textsize=tiny]{todonotes}
\usepackage{algorithm}
\let\aclAuthorAnd\AND
\let\AND\undefined
\usepackage{algorithmic}
\let\AND\aclAuthorAnd
\usepackage{enumitem}

\theoremstyle{plain}

\theoremstyle{definition}

\theoremstyle{remark}

\newcommand\our{\mbox{JURIST}}

\title{\our{}: Training Small Data-Reweighted Reasoning\texorpdfstring{\\}{ } Judges for Test-Time Scaling in Code Generation}

\author{Peijia Qin \quad Ruiyi Zhang \quad Qi Cao \quad Pengtao Xie \\
  University of California, San Diego}

\newcolumntype{Y}{>{\centering\arraybackslash}X}
\begin{document}
\maketitle
\ifdefined\dajpreprint
  \begingroup
  \let\thefootnote\relax
  \footnotetext{\parbox[t]{\dimexpr\columnwidth-1em\relax}{Correspondence to: Pengtao Xie $<$p1xie@ucsd.edu$>$. \\ Preprint. November 8, 2026.}}
  \endgroup
\fi
\let\AND\undefined

\begin{abstract}
  Best-of-N selection for code generation is only as good as the judge that picks among the candidates, and the judges accurate enough for this role are costly frontier LLMs. A small judge can be trained on abundant earlier data labeled by unit tests, but it is deployed on newer problems: \emph{the earlier problems are easier and teach it little, their tasks differ, and their candidates come from weaker policies}, while data close to deployment are scarce. We present \textbf{\our{}}, a 1.7B reasoning judge that uses a small meta set close to deployment to learn how much each earlier training sample should count. In our experiments, we find that training on the meta set alone, or adding it to the training data, does not help, while using it to weight the earlier data through a bi-level objective adds 3.9 Hard points over adding it to the training data. On LiveCodeBench, \our{} reaches 84.7\% Overall and \textbf{matches Claude Sonnet 4.6 on Hard problems (73.8\%) at 463$\times$ lower cost}; on BigCodeBench, it surpasses all test-time scaling baselines (35.9\%). Abundant earlier data, guided by a small deployment-aligned meta set, can thus train a compact judge that selects accurately at low cost.
\end{abstract}

\section{Introduction}
\begin{figure}[t]
  \centering
  \includegraphics[width=\linewidth]{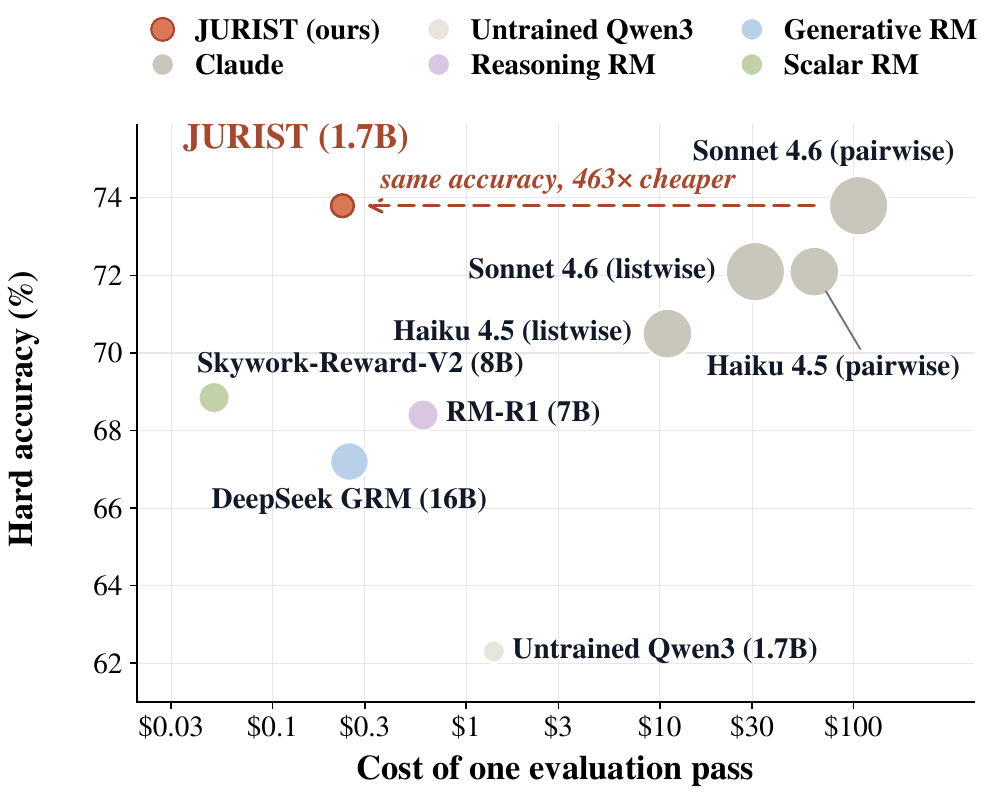}
  \caption{Cost-accuracy comparison among LLM judges on LiveCodeBench Hard problems. The x-axis is logarithmic; marker area grows with model size.}
  \label{fig:cost-accuracy}
\end{figure}
Test-time scaling (TTS) spends additional inference-time computation to improve the outputs of large language models (LLMs) on coding tasks~\citep{zhang2025survey,snell2025scaling,jaech2024openai,guo2025deepseekr1}. Selection-based TTS such as Best-of-N~\citep{irvine2023rewarding} samples multiple candidates and lets a judge select one, so the judge bounds the gain. LLM-as-a-Judge methods use LLMs themselves as this judge and reason about the candidates before selecting~\citep{zheng2023judging,wu2024meta}.

\begin{figure*}[t]
  \centering
  \includegraphics[width=\textwidth]{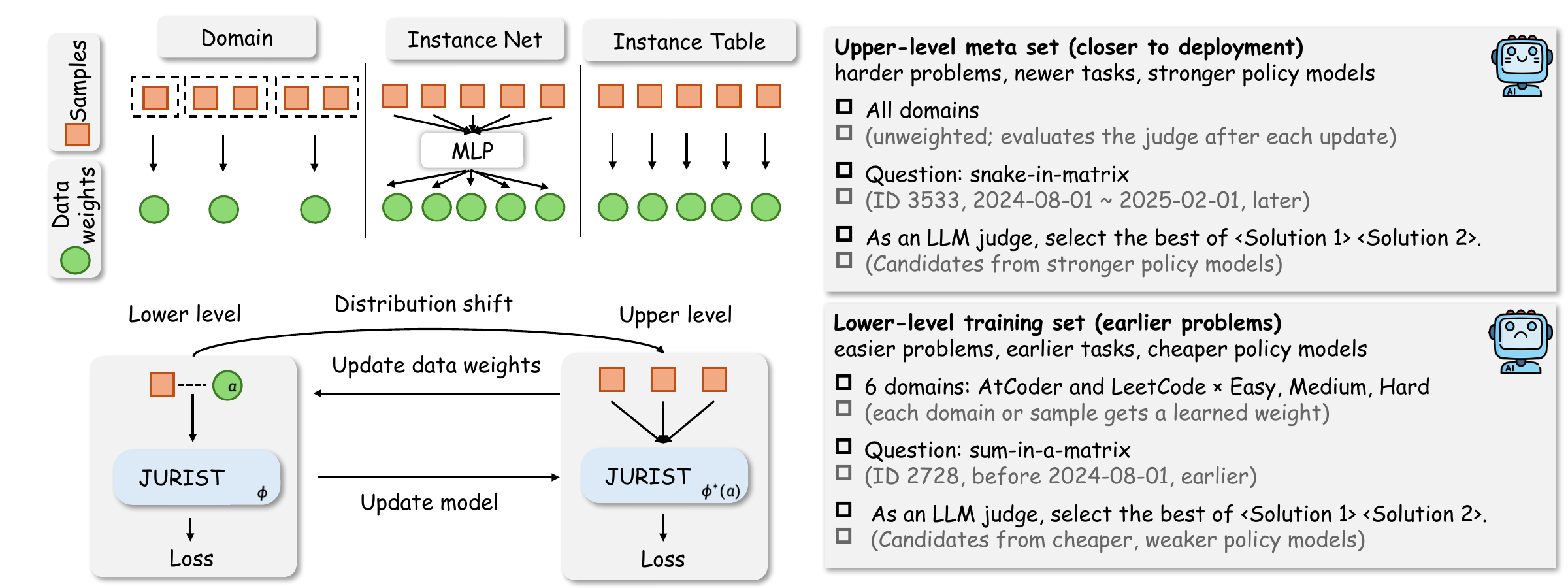}
  \caption{Bi-level data-reweighted training framework.}
  \label{fig:dream}
\end{figure*}

\begin{figure*}[t]
  \centering
  \includegraphics[width=\textwidth]{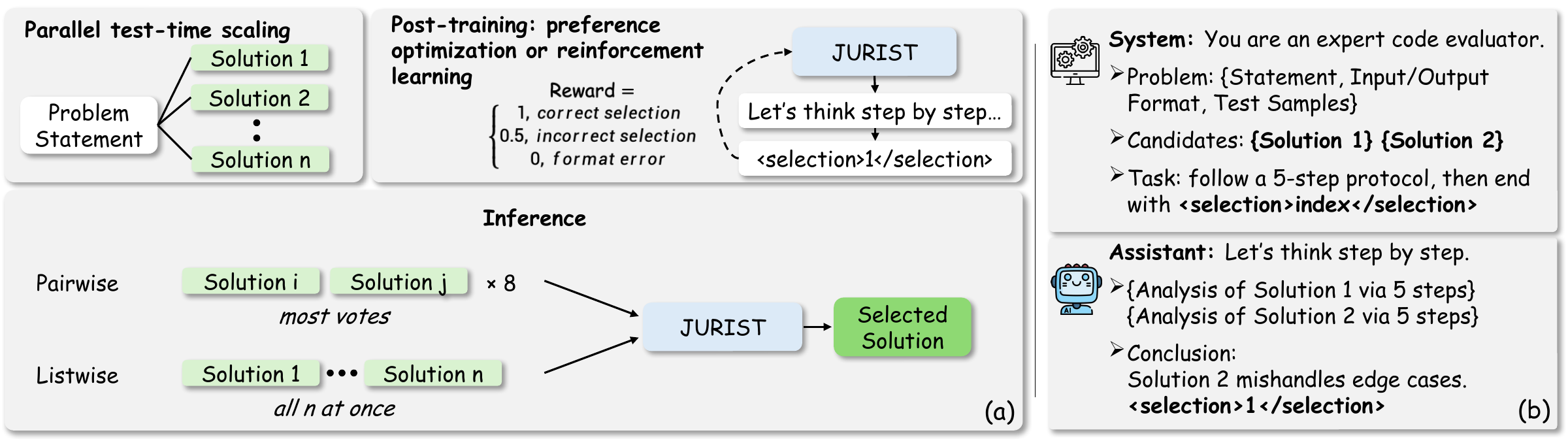}
  \caption{How \our{} is trained and how it selects a candidate at inference (a), with its prompt and output (b).}
  \label{fig:rl}
\end{figure*}

However, training an LLM judge for code generation is hard because \emph{its training data differ from the data it meets at deployment}. On LiveCodeBench, which releases new problems over time, a judge trained on the earlier problems is deployed on the later ones, and the two sets differ in three ways. The earlier problems are easier, and an untrained judge already solves many easy problems in every judgment, so they teach it little (\emph{difficulty mismatch}). Medium and hard problems still teach it (\Cref{app:operationalization}), and they make up 76.3\% of the 131 test problems. The training tasks also differ from the target tasks (\emph{task mismatch}), and the training candidates, the solutions and reasoning trajectories written by the \emph{policy models}, come from cheaper, weaker policy models than the test candidates (\emph{trajectory mismatch}).

We propose \our{}, a small \textit{\textbf{JU}dge from \textbf{R}easoning, \textbf{I}mportance reweighting, and \textbf{S}election \textbf{T}raining}, to address these shifts without discarding the earlier data (\Cref{fig:dream}). The key to an effective small judge is learning which training samples improve its decisions on the candidates it will face. \our{} keeps the earlier data and uses the few data close to deployment as a \emph{meta set} that learns a weight for each training sample; the meta set holds harder problems, tasks closer to the test-time tasks, and candidates from stronger models. These data are too few to train on: training on the meta set alone lowers Hard accuracy by 2.3 points and pooling it with the training data does not raise it, while using it to weight the training data raises it by 2.6 points (\Cref{sec:ablation-source}). A weight rises when the judge can still learn from the sample (its \emph{learning signal}) and learning from it lowers the loss on the meta set (the \emph{alignment} of its gradient with the meta-loss gradient; \Cref{sec:method}). The learned weights therefore \textbf{move training toward the problems that still teach the judge and address the three shifts at once}. In the middle of training, the weight rises for 62\% of medium and 60\% of hard training problems, against 45\% of easy ones (\Cref{app:operationalization}).

Code needs no annotation: unit tests tell whether the judge selects a passing candidate on every training and meta sample, and this one outcome defines both the judge's training loss and the meta loss (\Cref{fig:rl}). With 1.7B parameters, \our{} outperforms every test-time scaling baseline on LiveCodeBench and BigCodeBench (\Cref{tab:results}), and its bi-level training adds only 20.36\% wall-clock over single-level training, about 16 hours on $2\times$A100.

\begin{table*}[t]
  \centering
  \scriptsize
  \caption{Prior work on code Best-of-N selection against the four requirements of \Cref{app:related}. \pmark: holds in part.}
  \label{tab:related}
  \begin{tabularx}{\linewidth}{lYYYYYYY}
    \toprule
    & \multicolumn{2}{c}{(i) Explains} & \multicolumn{2}{c}{(ii) Execution} & \multicolumn{2}{c}{(iii) Shift} & (iv) Cost \\
    \cmidrule(lr){2-3}\cmidrule(lr){4-5}\cmidrule(lr){6-7}\cmidrule(lr){8-8}
    Method & Reasoning & Compares & \shortstack{Exec.\\labels} & \shortstack{Outcome\\RL} & Domain & Instance & Size \\
    \midrule
    \rowcolor{gray!20}\multicolumn{8}{c}{\textit{Generative judges}} \\
    Auto-J~\citep{li2024generative} & \cmark & \cmark & \xmark & \xmark & \xmark & \xmark & 13B \\
    Prometheus 2~\citep{kim2024prometheus} & \cmark & \cmark & \xmark & \xmark & \xmark & \xmark & 7B, 8$\times$7B \\
    Self-Taught Eval.~\citep{wang2024self} & \cmark & \cmark & \cmark & \xmark & \xmark & \xmark & 70B \\
    GenRM~\citep{zhang2025generative} & \cmark & \xmark & \pmark & \xmark & \xmark & \xmark & 2B--9B \\
    DeepSeek GRM~\citep{liu2025inference} & \cmark & \pmark & \xmark & \cmark & \xmark & \xmark & 16B--671B \\
    JudgeLRM~\citep{chen2025judgelrm} & \cmark & \cmark & \xmark & \cmark & \xmark & \xmark & 3B--14B \\
    J1~\citep{whitehouse2025j1} & \cmark & \cmark & \cmark & \cmark & \xmark & \xmark & 8B--70B \\
    RM-R1~\citep{chen2025rmr1} & \cmark & \cmark & \xmark & \cmark & \xmark & \xmark & 7B--32B \\
    \rowcolor{gray!20}\multicolumn{8}{c}{\textit{Scalar reward models}} \\
    PRM800K PRM~\citep{lightman2023verify} & \xmark & \xmark & \xmark & \xmark & \xmark & \xmark & GPT-4 \\
    Math-Shepherd~\citep{wang2023math} & \xmark & \xmark & \cmark & \xmark & \xmark & \xmark & 7B--70B \\
    FunPRM~\citep{zhang2026funprm} & \xmark & \xmark & \cmark & \xmark & \xmark & \pmark & 7B \\
    DreamPRM~\citep{cao2025dreamprm} & \xmark & \xmark & \cmark & \xmark & \cmark & \xmark & 2B \\
    DreamPRM-1.5~\citep{cao2025dreamprm15} & \xmark & \xmark & \cmark & \xmark & \xmark & \cmark & 1B \\
    \midrule
    \rowcolor{orange!12}\our{} (ours) & \cmark & \cmark & \cmark & \cmark & \cmark & \cmark & \textbf{1.7B, 14B} \\
    \bottomrule
  \end{tabularx}

\end{table*}

Our main contributions are:
\begin{itemize}[topsep=2pt, itemsep=3pt, parsep=0pt]
  \item \textbf{Data reweighting for generative judges.} We show that the training data of a generative judge can be steered toward deployment by weights learned against a small meta set, and that the meta gradient of each weight factors into a learning signal and an alignment.
  \item \textbf{A small judge that matches a frontier judge on Hard problems.} On LiveCodeBench, our 1.7B judge matches the Hard accuracy of Claude Sonnet 4.6 under the same pairwise protocol (73.8\%) at 463$\times$ lower cost (\Cref{fig:cost-accuracy}), and outperforms 16B and 7B reward models (\Cref{tab:results}).
  \item \textbf{Learned weights improve Hard selection under every objective.} Domain reweighting raises Hard accuracy under all four training objectives, and every reweighting strategy improves or matches the judge without reweighting (\Cref{tab:ablation,tab:grpo-ablation,tab:grpo-instance}). Training on the meta set alone, pooling it with the training set, fine-tuning on the strong-policy data, a domain filter, or fixed heuristic weights each fall 2.0--5.7 Hard points short (\Cref{sec:ablation-source,tab:robustness}).
\end{itemize}

\section{Related Work}
\label{sec:related}
\our{} is most related to three areas of prior work. \emph{LLM-as-a-Judge} methods write their reasoning before the verdict and are trained on preference data, on synthetic pairs, or by reinforcement learning against a reward that checks the verdict~\citep{zheng2023judging,li2024generative,kim2024prometheus,wang2024self,zhang2025generative,liu2025inference,chen2025judgelrm,whitehouse2025j1,chen2025rmr1}. \emph{Test-time scaling} spends more inference compute through longer reasoning or many sampled candidates, and outcome and process reward models score the candidates to select one~\citep{wei2022chain,guo2025deepseekr1,brown2024large,snell2025scaling,cobbe2021verifiers,lightman2023verify,wang2023math,he_2024_16998085,zhang2026funprm}. \emph{Data reweighting} learns how much each training sample or data domain contributes to the loss, for classifiers and for language-model pretraining~\citep{ren2018learning,shu2019meta,xie2023doremi,DOGE}. \Cref{tab:related} compares \our{} with these methods; among them, \our{} is a compact generative judge trained on unit-test outcomes whose training data are weighted toward deployment. See \Cref{app:related} for more detailed discussion of related work.

\section{Data-Reweighted Training of LLM Judges}
\label{sec:method}

\paragraph{Problem setup.}
\our{} is a reasoning-based LLM judge: given a problem and two candidate solutions, it writes its reasoning and then selects the preferred candidate (\Cref{fig:rl}(b); the prompt is in \Cref{appendix:prompts}). When more than two candidates are available, the judge runs $R$ rounds; each round compares a new random pair of candidates (all pairs when fewer than $R$ exist), and the preferred one receives a vote. The candidate with the most votes is selected, with ties broken at random. We denote the training data by $\mathcal{D}_{\text{tr}} = \{(X_i, Y_i)\}_{i=1}^{N}$, where $X_i$ is a problem paired with candidate solutions and $Y_i$ is the judge's written reasoning and its selection. Because each candidate can be executed against unit tests, the reward $\mathcal{R}(X_i, Y_i)$ is verifiable without human annotation: 1 for a correct selection, 0.5 for an incorrect one, and 0 for a format error, following the reinforcement learning with verifiable rewards paradigm~\citep{guo2025deepseekr1}. Unit tests serve only training and evaluation: at inference, the judge sees only the problem and the candidates. The judge parameters $\phi$ are trained either with GRPO on this reward or by preference optimization on pairs whose chosen response selects correctly and whose rejected one does not (\Cref{sec:preliminary}); we write $\mathcal{L}(X_i, Y_i; \phi)$ for the per-sample loss of either objective.

\paragraph{Bi-level data reweighting.}
\our{} assigns each training sample $(X_i, Y_i)$ a weight $\alpha_{(X_i, Y_i)}$ and learns the weights so that the judge trained on the weighted data performs well on a held-out meta set (\Cref{fig:dream}). The training set $\mathcal{D}_{\text{tr}}$ holds the many earlier or easier problems, with candidates from cost-efficient models; the smaller meta set $\mathcal{D}_{\text{meta}}$ holds problems closer to the test period or harder ones, with candidates from stronger policy models (\Cref{app:impl}). We parameterize the weights at three granularities (\Cref{tab:reweighting-methods}): \emph{domain reweighting} learns one weight $\alpha_k$ for each of $K$ domains (on LiveCodeBench, six: 2 platforms $\times$ 3 difficulty levels); \emph{instance table} stores one weight $\alpha_i$ per sample; and \emph{instance net} computes $\alpha_i = \mathrm{MLP}_\theta(\mathcal{L}(X_i, Y_i; \phi))$ from the sample's current loss, so its parameter count $|\theta|$ does not grow with the dataset size $N$. For the instance net, an update of $\alpha$ is an update of $\theta$.

\begin{table}[t]
  \centering
  \scriptsize
  \caption{Data-reweighting strategies.}
  \label{tab:reweighting-methods}
  {\setlength{\tabcolsep}{2.5pt}\begin{tabular}{lccl}
  \toprule
  Method          & Granularity & Params & Weight formula \\
  \midrule
  Domain reweighting          & domain  & $K$              & $\alpha_k$ assigned to domain $\mathcal{D}_k$ \\
  Instance table  & sample  & $N$              & $\alpha_i$ stored per $(X_i, Y_i)$ \\
  Instance net    & sample  & $|\theta|$       & $\alpha_i = \mathrm{MLP}_\theta(\mathcal{L}(X_i, Y_i; \phi))$ \\
  \bottomrule
  \end{tabular}}
\end{table}

\begin{table}[t]
  \centering
  \scriptsize
  \caption{Margin $m_i$ and learning signal $\psi_i$ of each objective; $q_i = \sigma(m_i)$, $p_i$ is the probability that the judge selects a passing candidate on sample $i$ (\Cref{app:operationalization}); $\lambda$ and $\lambda_{\text{O}}$ are the KTO class weight and the ORPO odds-ratio weight, respectively.}
  \label{tab:signal}
  \begin{tabularx}{\linewidth}{lXl}
    \toprule
    Objective & Margin $m_i$ & Learning signal $\psi_i$ \\
    \midrule
    DPO & log-ratio, chosen minus rejected & $1-q_i$ \\
    KTO & implicit reward minus reference & $\lambda\,q_i(1-q_i)$ \\
    ORPO & log odds ratio & $\lambda_{\text{O}}(1-q_i)$ \\
    GRPO & $\log p_i/(1-p_i)$ & $\sqrt{p_i(1-p_i)}$ \\
    \addlinespace[2pt]
    \bottomrule
  \end{tabularx}
\end{table}

\paragraph{Lower-level optimization.}
At the lower level, we train the judge parameters $\phi$ on the weighted training data while $\alpha$ stays fixed:
\begin{equation}
\phi^*(\alpha) = \arg\min_{\phi} \sum_{(X_i, Y_i)\in\mathcal{D}_{\text{tr}}} \alpha_{(X_i, Y_i)}\, \mathcal{L}(X_i, Y_i; \phi).
\end{equation}
We write the weighted sum as $\mathcal{L}_{\text{tr}}(\mathcal{D}_{\text{tr}}, \phi, \alpha)$. The trained judge $\phi^*$ is a function of $\alpha$.

\paragraph{Upper-level optimization.}
At the upper level, we update the data weights $\alpha$ so that the judge $\phi^*(\alpha)$ from the lower level minimizes the loss on the held-out meta set:
\begin{equation}
\label{eq:upper}
\alpha^* = \arg\min_{\alpha} \sum_{(X_i, Y_i)\in\mathcal{D}_{\text{meta}}} \mathcal{L}(X_i, Y_i; \phi^*(\alpha)).
\end{equation}
We write it as $\mathcal{L}_{\text{meta}}(\mathcal{D}_{\text{meta}}, \phi^*(\alpha))$. The meta samples carry no weights, so the meta loss measures the judge on data close to deployment.

\paragraph{Optimization algorithm.}
Solving \Cref{eq:upper} exactly requires training the judge to convergence for every update of $\alpha$, which is prohibitive. We approximate $\phi^*(\alpha)$ with one unrolled lower-level step, which meta-learning of sample weights already uses at scale~\citep{ren2018learning,shu2019meta,choe2023betty}. At iteration $t$, the judge first takes one gradient step on the weighted training loss with learning rate $\beta_1$:
\begin{equation}
\phi^{(t+1)} = \phi^{(t)} - \beta_1 \nabla_{\phi} \mathcal{L}_{\text{tr}}(\mathcal{D}_{\text{tr}}, \phi^{(t)}, \alpha^{(t)}).
\end{equation}
We then use $\phi^{(t+1)}$ in place of $\phi^*(\alpha)$ and update the weights with learning rate $\beta_2$, differentiating the meta loss through the lower-level step:
\begin{equation}
\alpha^{(t+1)} = \alpha^{(t)} - \beta_2 \nabla_{\alpha} \mathcal{L}_{\text{meta}}(\mathcal{D}_{\text{meta}}, \phi^{(t+1)}).
\end{equation}
The two steps alternate until convergence. \Cref{alg:daj} in \Cref{app:algorithm} gives the full procedure.

\paragraph{Discussion.}
Under the one-step unrolling, the upper-level gradient of each weight has a closed form. Let $P$ be the optimizer's preconditioner for the lower-level step ($P=I$ for SGD; the diagonal scaling for Adam), held fixed for the step, ignoring the optimizer's momentum. With the gradients
\begin{gather*}
g_i = \nabla_\phi \mathcal{L}(X_i,Y_i;\phi^{(t)}), \\ g_{\text{meta}} = \nabla_\phi \mathcal{L}_{\text{meta}}(\mathcal{D}_{\text{meta}},\phi^{(t+1)}),
\end{gather*}
the chain rule through $\phi^{(t+1)}$ gives
\begin{equation}
\label{eq:hypergrad}
-\frac{\partial \mathcal{L}_{\text{meta}}}{\partial \alpha_{(X_i,Y_i)}} = \beta_1 \big\langle g_i,\, P\, g_{\text{meta}} \big\rangle .
\end{equation}
DPO and KTO depend on $\phi$ through a scalar \emph{margin} $m_i$ that grows as the judge favors the correct judgment on sample $i$, and GRPO does so in expectation (\Cref{tab:signal}). Hence $\nabla_\phi \mathcal{L}(X_i,Y_i;\phi) = -\psi_i \nabla_\phi m_i$ with a \emph{learning signal} $\psi_i \ge 0$; ORPO adds a likelihood term that stays outside the product (\Cref{app:operationalization}). Substituting gives
\begin{equation}
\label{eq:factor}
-\frac{\partial \mathcal{L}_{\text{meta}}}{\partial \alpha_{(X_i,Y_i)}} = \beta_1\, \underbrace{\psi_i}_{\makebox[0pt]{$\substack{\text{signal}\\ \text{(difficulty)}}$}}\, \underbrace{\big\langle \nabla_\phi m_i,\, -P\, g_{\text{meta}} \big\rangle}_{\substack{\text{alignment}\\ \text{(task, trajectory)}}} .
\end{equation}
Since $\psi_i \ge 0$, the alignment sets the sign of the update and the learning signal scales it: a weight rises when the objective still moves the sample and raising its margin moves $\phi$ in the direction that lowers the meta loss.

\begin{figure*}[t]
  \centering
  \begin{minipage}[c]{0.5\textwidth}
    \centering
    \scriptsize
    \captionof{table}{The two factors of \Cref{eq:factor}: the shift each addresses, and whether domain reweighting (Domain), the instance table (Table) and the instance net (Net) can express it. \cmark: yes; \pmark: in part (\Cref{app:operationalization}).}
    \label{tab:factor-map}
    \begin{tabularx}{\linewidth}{l>{\raggedright\arraybackslash}Xccc}
      \toprule
      Factor & Shift it addresses & Domain & Table & Net \\
      \midrule
      Learning signal $\psi_i$ & Difficulty mismatch & \cmark & \cmark & \cmark \\
      Alignment & Task and trajectory mismatch; difficulty in part & \cmark & \cmark & \pmark \\
      \bottomrule
    \end{tabularx}
  \end{minipage}\hfill
  \begin{minipage}[c]{0.46\textwidth}
    \centering
    \includegraphics[width=\linewidth]{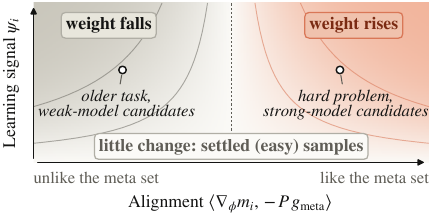}
    \caption{Sign and size of the update of \Cref{eq:factor} over its two factors. Curves: levels of their product.}
    \label{fig:update-map}
  \end{minipage}
\end{figure*}

\Cref{tab:factor-map} lists the shift each factor addresses and the weighting granularities that can express it, and \Cref{fig:update-map} shows how the two factors set the direction of the update. A sample whose margin is already large has a small $\psi_i$, so the learning signal moves weight away from settled samples, the easy problems that an untrained judge already solves in every judgment (\Cref{app:operationalization}). The meta set holds problems closer to the test period and candidates from stronger policy models, so the alignment separates samples that resemble the meta set from those that do not, which addresses task mismatch and trajectory mismatch. Both factors depend on $\phi$ and change as the judge trains, so learning the weights with the judge lets them follow (\Cref{sec:ablation-source}). \Cref{app:operationalization} gives the gradient analysis and an empirical study.

\begin{figure*}[t]
  \centering
  \includegraphics[width=\textwidth]{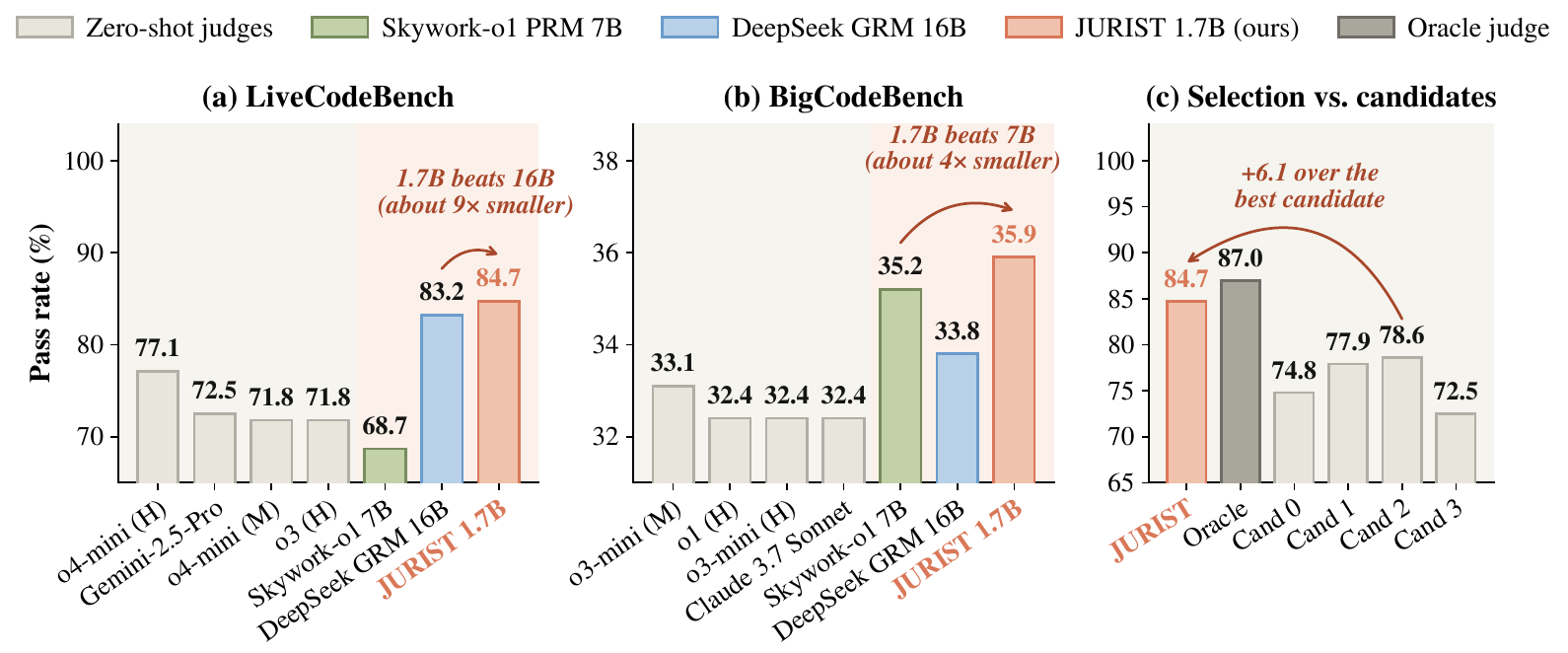}
  \caption{Overall pass@1 (\%) on (a) LiveCodeBench and (b) BigCodeBench: \our{} against zero-shot models that solve each problem directly (grey shading) and test-time scaling baselines; (H), (M): high, medium reasoning effort. (c) \our{} against each of the four DeepSeek V3.2 Speciale candidates and an oracle judge on LiveCodeBench. The y-axes do not start at zero.}
  \label{tab:results}
  \label{tab:results-bigcodebench}
  \label{fig:best}
\end{figure*}

\section{Results}
\label{sec:experiment}
\subsection{Experimental Settings}
\label{sec:impl-details}

We evaluate on LiveCodeBench~\citep{jain2025livecodebench} with a temporal split that mirrors deployment (training problems released before 2024-08-01, meta problems before 2025-02-01, and 131 test problems after; \Cref{app:datasets}) and on BigCodeBench~\citep{zhuo2024bigcodebench} with a difficulty-based split. The training set holds candidates from open models (Qwen3-Coder-30B-A3B and Qwen2.5-Coder-32B), and the meta set holds candidates from stronger policies (o4-mini (high) and DeepSeek V3.2 Speciale). We train Qwen3-1.7B~\citep{yang2025qwen3} with GRPO and Qwen2.5-Coder-14B~\citep{hui2024qwen2} with preference optimization. Baselines are zero-shot frontier models from the official leaderboards, random selection, Skywork-o1 PRM~\citep{he_2024_16998085}, and DeepSeek GRM~\citep{liu2025inference}. We report pass@1 (\%), the share of test problems whose selected candidate passes all unit tests; entries with $\pm$ give the mean and standard deviation over inference seeds, the sampling seeds of the judge at evaluation (seeds 42--51 in \Cref{tab:grpo-ablation}, 42--46 elsewhere); a caption marks any other five-seed mean, and every entry without $\pm$ comes from one evaluation pass. Unless noted otherwise, candidates come from DeepSeek V3.2 Speciale, and \our{} votes over $R{=}8$ rounds (\Cref{app:datasets,app:impl}).

\subsection{Main Results}
On LiveCodeBench (\Cref{tab:results}), \our{} reaches 84.7\% Overall, 7.6 points above the strongest zero-shot baseline, o4-mini (high), and 1.5 points above the strongest test-time scaling baseline, DeepSeek GRM 16B (83.2\%), with a backbone about $9\times$ smaller. On BigCodeBench, \our{} reaches 35.9\%, above the 35.2\% of Skywork-o1 PRM 7B. \Cref{appendix:per-difficulty-results} gives the per-difficulty breakdown.

\our{} is most accurate where selection matters. On the \emph{mixed-outcome} problems, which have both a passing and a failing candidate, \our{} selects a passing candidate in 90.2\% of cases, against 83.8\% for DeepSeek GRM and 53.2\% for random selection, and in 95.9\%, 79.1\%, and 47.9\% of the Hard ones (\Cref{tab:mixed-outcome}); on 63.4\% of the test problems every candidate passes. Against an oracle judge, which selects a correct candidate whenever one exists, \our{} closes 79\% of the gap between random selection (76.0\%) and the oracle (87.0\%) Overall and 92\% on Hard problems (54.9\% to 75.4\%), and it outperforms every individual candidate (\Cref{fig:best}(c)).

\begin{table}[t]
  \centering
  \scriptsize
  \caption{Overall accuracy across policy models on LiveCodeBench. o4-mini: o4-mini (high); V3.2: DeepSeek V3.2 Speciale; Qwen2.5: Qwen2.5-Coder-32B; Qwen3-30B: Qwen3-Coder-30B-A3B.}
  \label{tab:results-policy}
  \begin{tabular}{lccccc}
      \toprule
      Method & o4-mini & V3.2 & Qwen2.5 & Qwen3-30B & Avg. \\
      \midrule
      Skywork-o1 PRM 7B  & 74.1          & 68.7          & 25.2          & 36.6          & 51.2          \\
      DeepSeek GRM 16B   & 75.6          & 83.2          & 25.2          & 38.2          & 55.6          \\
      \rowcolor{orange!12}\our{} 1.7B (ours) & \textbf{81.7} & \textbf{84.7} & \textbf{26.7} & \textbf{38.9} & \textbf{58.0} \\
      \bottomrule
    \end{tabular}
  \vspace{12pt}
  \caption{Accuracy on all and on mixed-outcome LiveCodeBench problems, which have passing and failing candidates (DeepSeek V3.2 Speciale candidates).}
  \label{tab:mixed-outcome}
  \begin{tabular}{lcccc}
    \toprule
    & \multicolumn{2}{c}{Overall} & \multicolumn{2}{c}{Hard} \\
    \cmidrule(lr){2-3}\cmidrule(lr){4-5}
    Method & All & Mixed-outcome & All & Mixed-outcome \\
    \midrule
    Random & 76.0 & 53.2 & 54.9 & 47.9 \\
    DeepSeek GRM 16B & 83.2 & 83.8 & 67.2 & 79.1 \\
    \rowcolor{orange!12}\our{} 1.7B (ours) & \textbf{84.7} & \textbf{90.2} & \textbf{73.8} & \textbf{95.9} \\
    \bottomrule
  \end{tabular}
\end{table}

\paragraph{Comparison with frontier judges.}
\label{sec:frontier-judges}
We run Claude Sonnet 4.6 and Haiku 4.5 under the pairwise voting protocol of \our{}, on the same LiveCodeBench problems and DeepSeek V3.2 Speciale candidates, and report their listwise results, one judgment over all candidates of a problem. Cost is in USD per evaluation pass and covers judging only, not candidate generation: the Claude judges at API prices, and \our{} at the A100 80GB list rate (\$2/hour\protect\footnote{Lambda Labs on-demand A100 80GB cloud rental rate (\url{https://lambdalabs.com}).}).

Under pairwise voting, Claude Sonnet 4.6 and Haiku 4.5 reach 86.3\% and 85.5\% Overall, against 84.7\% for \our{} (\Cref{fig:cost-accuracy}; exact values in \Cref{tab:frontier-judges}); on Hard problems Sonnet 4.6 matches \our{} at 73.8\% and Haiku 4.5 reaches 72.1\%. They cost \$62.85 to \$106.48 per pass, 273 to 463 times the \$0.23 of \our{}. Against four public reward models (1.7B--8B; \Cref{app:public-rms}), averaged over four policy models, \our{} has the highest Overall and Hard accuracy (\Cref{tab:public-rms}).

\subsection{Results on Different Policy Models}
With a 1.7B backbone, \our{} outperforms both baselines in Overall pass@1 for each of four policy models, o4-mini (high)~\citep{openai_o3_o4mini_system_card}, DeepSeek V3.2 Speciale~\citep{deepseekai2025deepseekv32}, Qwen2.5-Coder-32B~\citep{hui2024qwen2}, and Qwen3-Coder-30B-A3B~\citep{yang2025qwen3} (\Cref{tab:results-policy}), although they have 7B (Skywork-o1 PRM) and 16B (DeepSeek GRM) parameters. Averaged over the four policy models, \our{} reaches 58.0\% Overall, against 51.2\% for Skywork-o1 PRM and 55.6\% for DeepSeek GRM.

\subsection{Reweighting across Training Objectives}
\label{sec:ablation-objectives}
Learned weights improve the judge under every training objective we test. We compare four strategies (no reweighting, domain reweighting, instance table, and instance net) under the preference-optimization objectives direct preference optimization (DPO)~\citep{rafailov2023direct}, KTO~\citep{ethayarajh2024kto}, and ORPO~\citep{hong2024orpo} and under GRPO~\citep{guo2025deepseekr1}; the objectives differ only in the margin and learning signal of \Cref{eq:factor} (\Cref{tab:signal}).

\paragraph{Preference optimization.}
Across DPO, KTO, and ORPO, every reweighting strategy outperforms or matches no reweighting on Overall and Hard accuracy (\Cref{tab:ablation}), with the largest gains on Hard problems. Averaged over the three objectives, instance net reaches the highest Overall accuracy and instance table the highest Hard accuracy (\Cref{tab:ablation}).

\begin{table}[t]
  \centering
  \scriptsize
  \caption{Reweighting under three preference-optimization objectives on LiveCodeBench; all difficulty levels in \Cref{tab:ablation-full}.}
  \label{tab:ablation}
  \begin{tabularx}{\linewidth}{lYYYY}
  \toprule
  Method & DPO & KTO & ORPO & Average \\
  \midrule
  \rowcolor{gray!20}\multicolumn{5}{c}{\textit{Overall}} \\
  No reweighting & 81.7 & 83.2 & 81.7 & 82.2 \\
  Domain reweighting & \textbf{83.2} & 83.2 & 82.4 & 82.9 \\
  Instance table & 82.4 & \textbf{84.0} & \textbf{83.2} & 83.2 \\
  Instance net & \textbf{83.2} & \textbf{84.0} & \textbf{83.2} & \textbf{83.5} \\
  \midrule
  \rowcolor{gray!20}\multicolumn{5}{c}{\textit{Hard}} \\
  No reweighting & 65.6 & 67.2 & 65.6 & 66.1 \\
  Domain reweighting & 67.2 & 68.9 & 68.9 & 68.3 \\
  Instance table & 65.6 & \textbf{70.5} & \textbf{70.5} & \textbf{68.9} \\
  Instance net & \textbf{68.9} & 68.9 & 67.2 & 68.3 \\
  \bottomrule
  \end{tabularx}
\end{table}

\paragraph{GRPO.}
Over ten inference seeds, domain reweighting outperforms no reweighting by 1.37 points Overall and 2.62 points on Hard problems, and both gains are significant under a paired $t$-test ($p{=}0.027$ and $p{=}0.008$; \Cref{tab:grpo-ablation}); the instance table and the instance net also improve on no reweighting (\Cref{app:grpo-instance}).

\begin{table}[t]
  \centering
  \scriptsize
  \caption{GRPO ablation on LiveCodeBench, mean $\pm$ standard deviation over ten inference seeds; $p$: paired $t$-test against domain reweighting over the seeds.}
  \label{tab:grpo-ablation}
  \setlength{\tabcolsep}{2.5pt}
  \begin{tabular*}{\linewidth}{@{}l@{\extracolsep{\fill}}cccc@{}}
      \toprule
      Method & Overall & $p$ & Hard & $p$ \\
      \midrule
      \rowcolor{gray!20}\multicolumn{5}{c}{\textit{Reweighting strategy}} \\
      No reweighting & 83.36 $\pm$ 1.29 & .027 & 69.02 $\pm$ 2.25 & .008 \\
      Domain reweighting & \textbf{84.73 $\pm$ 0.80} & -- & \textbf{71.64 $\pm$ 1.11} & -- \\
      \midrule
      \rowcolor{gray!20}\multicolumn{5}{c}{\textit{Single-level alternatives}} \\
      Meta set only & 82.14 $\pm$ 1.73 & .001 & 66.72 $\pm$ 3.00 & .002 \\
      Pooled training and meta sets & 82.82 $\pm$ 0.83 & .001 & 67.70 $\pm$ 1.90 & .002 \\
      Strong-policy SFT & 82.06 $\pm$ 1.09 & $<$.001 & 65.90 $\pm$ 2.54 & $<$.001 \\
      Domain filter & 82.21 $\pm$ 1.44 & .001 & 66.88 $\pm$ 2.42 & $<$.001 \\
      \bottomrule
    \end{tabular*}
  \vspace{12pt}
  \caption{Ablation of the meta set, reweighting strategy and inference reasoning. Four-policy avg: mean over the four policy models; single-policy: candidates of DeepSeek V3.2 Speciale.}
  \label{tab:robustness}
  \begin{tabularx}{\linewidth}{lYY}
      \toprule
      Method & Overall & Hard \\
      \midrule
      \rowcolor{gray!20}\multicolumn{3}{c}{\textit{Meta-set composition (four-policy avg)}} \\
      BigCodeBench meta set         & 57.4           & \textbf{38.9} \\
      LiveCodeBench meta set               & \textbf{58.0}  & \textbf{38.9} \\
      \midrule
      \rowcolor{gray!20}\multicolumn{3}{c}{\textit{Reweighting strategy (four-policy avg)}} \\
      Fixed inverse-frequency weights         & 56.5           & 36.9 \\
      Domain reweighting        & \textbf{58.0}  & \textbf{38.9} \\
      \midrule
      \rowcolor{gray!20}\multicolumn{3}{c}{\textit{Inference reasoning (single-policy)}} \\
      Direct selection          & 83.2           & 70.5 \\
      Chain-of-thought reasoning   & \textbf{84.7}  & \textbf{73.8} \\
      \bottomrule
    \end{tabularx}
\end{table}

\begin{figure}[t]
  \centering
  \includegraphics[width=\linewidth]{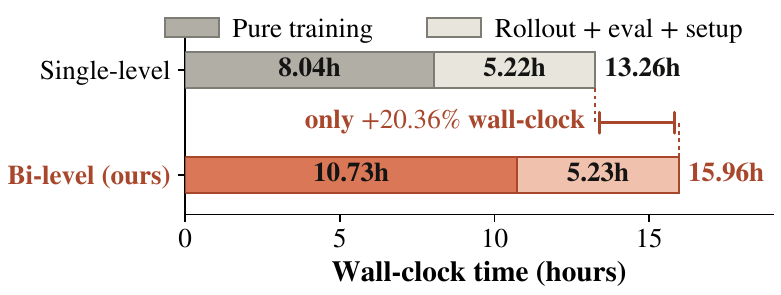}
  \caption{Training time on 2$\times$A100; the upper level adds 20.36\% wall-clock.}
  \label{fig:training-cost}
\end{figure}

\subsection{Ablation Study}
\label{sec:ablation-source}
\label{sec:ablation-robustness}
We investigate how a small meta set close to deployment is best used to train a judge; of the uses we compare, only weighting the training data against it helps. Each ablation changes one component of the proposed setting; \Cref{app:impl} defines each one, and \Cref{app:voting-rounds} varies the number of voting rounds.

\paragraph{Dataset use.}
Of four ways to use the two sets, training on the training set alone, on the meta set alone, on both pooled, or bi-level (\Cref{tab:grpo-ablation}), training on the meta set alone gives the lowest accuracy (82.14\% Overall, 66.72\% Hard); the meta set is small. Pooling the two sets does not exceed training on the training set alone (82.82\% against 83.36\% Overall; 67.70\% against 69.02\% Hard). Domain reweighting reaches 84.73\% and 71.64\%, 1.91 and 3.94 points above pooling, so the gain comes from using the meta set to steer the training data, not from training on it.

\paragraph{Strong-policy supervision.}
Supervised fine-tuning (SFT) of the judge directly on all 442 pairs built from the strong policy models that supply the meta set (strong-policy SFT), instead of using them as the meta set, reaches 82.06\% Overall and 65.90\% Hard, 2.67 and 5.74 points below domain reweighting (\Cref{tab:grpo-ablation}): these data help more as the upper-level objective than as training data.

\paragraph{Domain filter.}
A domain filter that drops training samples until the rest match the domain composition of the meta set keeps 42.7\% of the training set (\Cref{tab:grpo-ablation}). It does not improve on the unfiltered training set and falls 2.52 points Overall and 4.76 points Hard below domain reweighting, since a 0/1 rule can downweight a domain only by discarding its samples.

\paragraph{Fixed heuristic weights.}
Fixed inverse-frequency weights over the six domains (3 difficulty levels $\times$ 2 platforms), each proportional to $1/(\text{domain size})$ and normalized to sum to one, lower the four-policy averages by 1.5 points Overall and 2.0 points Hard relative to learned weights (\Cref{tab:robustness}). Both terms of \Cref{eq:factor} change as the judge trains, and neither fixed rule can follow them.

\paragraph{Meta-set composition.}
With a BigCodeBench meta set, the four-policy averages drop from 58.0\% to 57.4\% Overall and stay at 38.9\% on Hard, so \our{} loses little with a meta set from another benchmark.

\paragraph{Inference reasoning.}
With direct selection, the judge emits its selection without chain-of-thought reasoning, and accuracy drops by 1.5 points Overall and 3.3 points Hard (\Cref{tab:robustness}).

\paragraph{Listwise judging.}
With $R{=}8$, a pairwise judge compares every pair of a four-candidate problem (6 pairs; o4-mini and DeepSeek V3.2 Speciale) but fewer than a third of the 28 pairs of an eight-candidate problem (Qwen2.5-Coder-32B and Qwen3-Coder-30B). A listwise judge reads all candidates of a problem and selects one in a single judgment; we train a listwise variant of \our{} with the same bi-level framework.

Averaged over four policy models and five seeds (\Cref{tab:listwise}), the listwise judge reaches 57.7\% Overall and 37.1\% Hard with domain weights, and 57.7\% and 36.7\% with an instance weight table. With one judgment per problem, the domain-weighted judge exceeds DeepSeek GRM, which also judges once, by 2.1 and 1.8 points, and RM-R1, which casts up to eight pairwise votes, by 1.2 and 0.9 points. It uses one seventh of the pairwise judgments (524 vs.\ 3{,}668).

\begin{table}[t]
  \centering
  \scriptsize
  \caption{Inference cost on LiveCodeBench. Tokens per judgment: mean (range) over four policy models; Minutes (one evaluation pass), Overall, and Hard: DeepSeek V3.2 Speciale candidates. All entries are means over five inference seeds.}
  \label{tab:inference-cost}
  \begin{tabular}{lcccc}
      \toprule
      Judge & Tokens per judgment & Minutes & Overall & Hard \\
      \midrule
      Untrained Qwen3-1.7B & 5.7k (5.3k--6.4k) & 42 & 79.85 & 61.97 \\
      RM-R1 7B & 4.0k (3.1k--4.4k) & 18 & 82.52 & 68.31 \\
      \rowcolor{orange!12}\our{} 1.7B (ours) & \textbf{1.2k (1.2k--1.3k)} & \textbf{7} & \textbf{84.43} & \textbf{71.48} \\
      \bottomrule
    \end{tabular}
  \vspace{12pt}
  \caption{Comparison with other judges on LiveCodeBench, averaged over four policy models. Judgments: judge calls over 131 problems $\times$ 4 policy models. RM-R1 and the two listwise rows: means over five inference seeds.}
  \label{tab:listwise}
  \label{tab:public-rms}
  {\setlength{\tabcolsep}{2.4pt}\begin{tabular}{llcccc}
    \toprule
    Judge & Type & Size & Judgments & Overall & Hard \\
    \midrule
    Random & -- & -- & -- & 53.0 & 32.5 \\
    AceCodeRM & Scalar, code & 7B & -- & 54.2 & 32.4 \\
    Skywork-Reward-V2 & Scalar & 8B & -- & 55.1 & 34.4 \\
    Skywork-Reward-V2 & Scalar & 1.7B & -- & 56.9 & 36.7 \\
    DeepSeek GRM & Generative & 16B & 524 & 55.6 & 35.3 \\
    RM-R1 & Pairwise & 7B & 3{,}668 & 56.5 & 36.2 \\
    \midrule
    \our{}, domain & Listwise & 1.7B & 524 & 57.7 & 37.1 \\
    \our{}, instance table & Listwise & 1.7B & 524 & 57.7 & 36.7 \\
    \rowcolor{orange!12}\our{} & Pairwise ($R{=}8$) & 1.7B & 3{,}668 & \textbf{58.0} & \textbf{38.9} \\
    \bottomrule
  \end{tabular}}
\end{table}

\subsection{Training and Inference Cost}
\label{sec:training-cost}
\label{sec:inference-cost}
Bi-level reweighting adds a one-time training cost and no inference cost. On 2$\times$A100 80GB over 20{,}000 iterations, bi-level training takes 15.96h against 13.26h for single-level training (+20.36\%), and the upper-level pass lengthens only the training portion (\Cref{fig:training-cost}). At inference, \our{} is both more accurate and cheaper than RM-R1 and the untrained Qwen3-1.7B (\Cref{tab:inference-cost}). All three judges reason before a pairwise verdict, with $R{=}8$, on one A100 80GB. \our{} writes 1.2k tokens per judgment, against 4.0k and 5.7k, and one evaluation pass takes 7 minutes, against 18 and 42.

\section{Conclusion}
We presented \our{}, a 1.7B data-reweighted reasoning judge for Best-of-N code selection. \our{} learns training-data weights with a bi-level objective against a small meta set that is closer to deployment, which addresses the three mismatches at once. On LiveCodeBench, the small judge matches Claude Sonnet 4.6 on Hard problems (73.8\%) at 463$\times$ lower cost, and reweighting improves the judge under every training objective we test.
\section*{Limitations}
While this work studies code selection, where every candidate can be executed against unit tests so that the reward, the training pairs and the meta set all come from execution results without human or LLM annotation, the bi-level reweighting itself only requires an outcome that can be checked automatically. The same framework could extend to other such domains, such as mathematical reasoning with verifiable answers, and to benchmarks beyond LiveCodeBench and BigCodeBench.

Our experiments also focus on small judges, Qwen3-1.7B and Qwen2.5-Coder-14B, because a judge that is called many times for every problem must stay cheap. Although this choice keeps \our{} practical, larger judges remain a natural next step for data reweighting: stronger base judges leave more easy samples without a learning signal (\Cref{fig:nosignal-scale}).

\bibliography{main}

\appendix
\raggedbottom
\crefalias{section}{appendix}

\section{More Related Work}
\label{app:related}
The columns of \Cref{tab:related} are defined as follows. \emph{Reasoning}: writes its reasoning before its verdict. \emph{Compares}: one input holds two or more candidates. \emph{Exec.\ labels}: trained only on known answers, unit tests, or construction, with no human or LLM annotation. \emph{Outcome RL}: trained by reinforcement learning or preference optimization against an outcome reward. \emph{Domain}, \emph{Instance}: learns training-data weights per pool or per sample. \emph{Size}: parameters of the reported judges.

A judge for code Best-of-N selection should meet four requirements. It should (i) explain each verdict, so that a wrong selection can be traced, and compare the candidates against each other; (ii) learn from execution results, which need no human or LLM labels; (iii) stay accurate when the tasks and the policy that writes the candidates differ from those seen in training; and (iv) be small enough to run on every problem. \Cref{tab:related} checks generative judges and scalar reward models against them. Earlier generative judges meet (i) and some meet (ii), but none learns how much each training sample counts; earlier reward models that learn data weights output a scalar score and give no reasoning. \our{} meets all four.

\paragraph{LLM-as-a-Judge.}
LLM judges evaluate open-ended outputs in natural language, and their agreement with human preferences made them a standard evaluator~\citep{zheng2023judging}. Generative judges write their reasoning before the verdict. Auto-J and Prometheus 2 are fine-tuned on preference data and on judgments written by GPT-4~\citep{li2024generative,kim2024prometheus}; the Self-Taught Evaluator is fine-tuned iteratively on synthetic pairs whose better response is known by construction~\citep{wang2024self}; and GenRM verifies one solution at a time by next-token prediction~\citep{zhang2025generative}. A more recent line trains the judge by reinforcement learning against a reward that checks its verdict: DeepSeek GRM~\citep{liu2025inference}, JudgeLRM~\citep{chen2025judgelrm}, J1~\citep{whitehouse2025j1}, and RM-R1~\citep{chen2025rmr1}. All of them train on their data as it is, with every sample counted equally, and none targets code selection. \our{} keeps the generative form and the outcome reward, takes its reward from unit tests, and learns weights that move the training distribution toward the deployment distribution: harder problems, the target tasks, and candidates from stronger policies.

\paragraph{Test-time scaling and reward models.}
Test-time scaling spends more inference compute to get a better output, either sequentially, by longer reasoning~\citep{wei2022chain,guo2025deepseekr1}, or in parallel, by sampling many candidates and aggregating them~\citep{brown2024large,snell2025scaling}. Parallel scaling needs a judge to pick the answer. Outcome reward models score a complete response~\citep{cobbe2021verifiers}, and process reward models score each reasoning step, trained on human step labels~\citep{lightman2023verify} or on Monte Carlo estimates checked against the answer~\citep{wang2023math}. For code, Skywork-o1 PRM scores code and mathematics steps~\citep{he_2024_16998085}, and FunPRM treats each function as a step~\citep{zhang2026funprm}. These models output a scalar per candidate and give no reasoning. \our{} compares two candidates in writing and returns a selection: complexity, edge cases, and algorithmic soundness are argued, not scored.

\paragraph{Data reweighting.}
Data reweighting changes how much each training sample contributes to the loss. Learning to reweight examples sets each sample weight from the inner product of its gradient with the gradient of a validation loss~\citep{ren2018learning}. Meta-Weight-Net learns a sample-level weighting function for classifiers trained on noisy or imbalanced labels~\citep{shu2019meta}, and DoReMi and DOGE learn domain-level mixture weights for language-model pretraining~\citep{xie2023doremi,DOGE}. \our{} learns such weights for a generative judge (\Cref{tab:related}): the model compares two programs and writes its reasoning, and the lower level trains it by reinforcement learning or preference optimization against an outcome reward from unit tests. Both domain-level and instance-level weights fit in one framework, and the weights target the three shifts between judge training and deployment. For the judge objectives, we show that this inner product factors into a learning signal and an alignment term (\Cref{sec:method}).

\section{Judging Task and Training Objectives}
\label{sec:preliminary}
\paragraph{Reasoning-based judging.}
\our{} belongs to the LLM-as-a-Judge paradigm: it writes explicit reasoning before its judgment~\citep{liu2025inference,ye2025learning}. Assessing code calls for such reasoning: time and space complexity, correctness on edge cases, and algorithmic soundness.

As illustrated in \Cref{fig:rl}(b), the judging task is formulated as a chat completion problem. Given a query and the corresponding candidate responses, \our{} autoregressively generates a thinking process followed by a final judgment that identifies the preferred candidate.
We use pairwise comparison, which leaves the output length for reasoning. When more than two candidates are available, we repeat the following for $R$ rounds: a pair of candidates that no earlier round has compared is drawn uniformly at random, the judge compares them, and the preferred candidate receives one vote; when a problem has fewer than $R$ pairs, every pair is compared once (6 rounds for four candidates). After all rounds, the candidate with the highest cumulative vote count is selected as the final output, with ties broken at random.
The full prompt template used for reasoning-based judging is provided in \Cref{appendix:prompts}.

\paragraph{Verifiable rewards.}
We consider a dataset $\mathcal{D}_{\text{tr}} = \{(X_i, Y_i)\}_{i=1}^{N}$, where $X_i$ denotes a problem statement paired with candidate solutions, and $Y_i$ represents the judge's selection along with a generated reasoning trajectory. Because each candidate can be executed against test cases, selection correctness is automatically verifiable without human annotations. We define the reward function as:
\begin{equation}
\label{eq:reward}
\mathcal{R}(X_i, Y_i) =
\begin{cases}
1, & \text{\our{} selects correctly} \\
0.5, & \text{\our{} selects incorrectly} \\
0, & \text{format error}
\end{cases}
\end{equation}
The reward $\mathcal{R}$ evaluates selection correctness rather than the quality of the candidate solutions themselves. This formulation follows the reinforcement learning with verifiable rewards (RLVR) paradigm~\citep{guo2025deepseekr1}, in which rule-based rewards supervise the model to develop reasoning patterns that lead to correct final selections, without requiring labeled reasoning traces.

\paragraph{Training objectives.}
Building on the above verifiable reward, we consider two training formulations. We train with Group Relative Policy Optimization (GRPO), which gives the loss $\mathcal{L}_{\text{RL}}(X_i, Y_i; \phi)$, where $\phi$ denotes the parameters of the judge model.
As an alternative, we cast judge training as a preference optimization problem. Preference pairs are constructed by treating correct selections as chosen responses and incorrect selections or format errors equally as rejected responses, collapsing the three-level reward into a binary preference signal. For each sample, the chosen judgment $y_i^{+}$ selects correctly and the rejected judgment $y_i^{-}$ selects incorrectly or has a format error. The loss is $\mathcal{L}_{\text{PO}}(X_i, Y_i; \phi)$.
For simplicity, we use $\mathcal{L}(X_i, Y_i; \phi)$ to denote the training loss, encompassing both formulations.

\section{Training Algorithm}
\label{app:algorithm}
\Cref{alg:daj} gives the full training procedure of \our{}: each iteration takes one lower-level step on the judge parameters $\phi$ and one upper-level step on the data weights $\alpha$, as described in \Cref{sec:method}.

\begin{algorithm*}[htbp]
  \caption{Data-Reweighted LLM Judge (\our{}).}
  \small
  \label{alg:daj}
  \begin{algorithmic}[1]
    \REQUIRE Training set $\mathcal{D}_{\text{tr}}$, meta set $\mathcal{D}_{\text{meta}}$, learning rates $\beta_1$, $\beta_2$
    \ENSURE Optimized judge parameters $\phi^*$, data weights $\alpha^*$
    \STATE $\phi^{(0)} \leftarrow$ pretrained LLM checkpoint; \ $\alpha^{(0)} \leftarrow 1/K$ per domain, $1$ per sample, or a random instance net (\Cref{tab:reweighting-methods})
    \WHILE{not converged}
      \STATE $\phi^{(t+1)} = \phi^{(t)} - \beta_1 \nabla_{\phi} \mathcal{L}_{\text{tr}}(\mathcal{D}_{\text{tr}}, \phi^{(t)}, \alpha^{(t)})$ \hfill $\triangleright$ Lower level: train the judge on the weighted data
      \STATE $\alpha^{(t+1)} = \alpha^{(t)} - \beta_2 \nabla_{\alpha} \mathcal{L}_{\text{meta}}(\mathcal{D}_{\text{meta}}, \phi^{(t+1)})$ \hfill $\triangleright$ Upper level: update the weights on the meta loss
    \ENDWHILE
    \STATE \textbf{return} $\phi^*$, $\alpha^*$
  \end{algorithmic}
\end{algorithm*}

\section{Datasets and Benchmarks}
\label{app:datasets}
\paragraph{LiveCodeBench.}
LiveCodeBench~\citep{jain2025livecodebench} is a benchmark designed to evaluate large language models on code-related tasks. It continuously collects problems from LeetCode, AtCoder, and Codeforces, each annotated with release dates to enable temporal evaluation. The benchmark covers four coding scenarios: code generation, self-repair, code execution, and test output prediction, with problems categorized into easy, medium, and hard difficulty levels. We use the code-generation task, excluding Codeforces problems due to insufficient samples. In our setup, 601 problems published before 2025-02-01 are used for training, while 131 problems published between 2025-02-01 and 2025-05-01 are reserved for testing. We further construct a hierarchical training split, in which the training set comprises problems released before 2024-08-01, and the meta set comprises problems released between 2024-08-01 and 2025-02-01, thereby providing stronger alignment with the target test distribution. Because the benchmark keeps adding problems, this split mirrors a judge that is trained on past problems and deployed on new ones; it separates the training, meta and test problems by release date, so the meta set problems are released before the test period and are disjoint from the test problems.

\paragraph{BigCodeBench.}
BigCodeBench~\citep{zhuo2024bigcodebench} evaluates LLMs on practical programming tasks that require tool use and complex instruction following. The benchmark contains 1{,}140 function-level tasks involving the composition of multiple function calls drawn from 139 libraries. Each task is associated with an average of 5.6 test cases and achieves approximately 99\% branch coverage. BigCodeBench includes two benchmark variants: BigCodeBench-Complete for code completion and BigCodeBench-Instruct for instruction following. BigCodeBench-Hard is a subset of BigCodeBench comprising 148 particularly challenging real-world tasks.
In our experiments, we evaluate on BigCodeBench-Instruct using the hard split, while the remaining tasks are used for training.
We split the training data into two subsets based on difficulty, measured by the pass rates of four policy models (o4-mini (high), DeepSeek V3.2 Speciale, Qwen2.5-Coder-32B, and Qwen3-Coder-30B-A3B).
The hardest 20\% of the tasks are used for the meta set, and the remaining 80\% are used for the training set. This is the same situation without release dates: plentiful training tasks, and a small set of harder tasks as the meta set.

\paragraph{Data domains.}
On LiveCodeBench, we define six domains based on the cross product of platform and difficulty level: AtCoder-easy, AtCoder-medium, AtCoder-hard, LeetCode-easy, LeetCode-medium, and LeetCode-hard. For BigCodeBench, we define seven domains by task category: Computation, Visualization, System, Time, Network, Cryptography, and General.

\section{Implementation Details}
\label{app:impl}
\paragraph{Models.}
Within these benchmark-specific splits, we construct the training set using reasoning trajectories and candidate solutions from Qwen3-Coder-30B-A3B~\citep{yang2025qwen3} and Qwen2.5-Coder-32B~\citep{hui2024qwen2}.
For the upper-level optimization, we employ o4-mini (high)~\citep{openai_o3_o4mini_system_card} and DeepSeek V3.2 Speciale~\citep{deepseekai2025deepseekv32} to provide higher-quality data. The meta set therefore lies closer to the test distribution than the training set.
We fine-tune Qwen2.5-Coder-14B~\citep{hui2024qwen2} with preference optimization (DPO, KTO and ORPO) and Qwen3-1.7B~\citep{yang2025qwen3} with GRPO.

\paragraph{Baselines.}
For evaluation, we compare \our{} against two categories of baselines: zero-shot methods and test-time scaling approaches. Zero-shot baselines are leading LLMs without test-time scaling, such as o4-mini and Gemini-2.5-Pro. Reported results are taken directly from the official LiveCodeBench and BigCodeBench leaderboards. We additionally compare against representative test-time scaling baselines. \textbf{Random} selects a candidate solution uniformly at random, reflecting the average performance of sampled candidates. \textbf{Skywork-o1 PRM}~\citep{he_2024_16998085} is a process reward model that assigns step-level scores during generation\footnote{\url{https://huggingface.co/Skywork/Skywork-o1-Open-PRM-Qwen-2.5-7B}}. \textbf{DeepSeek GRM}~\citep{liu2025inference} serves as a generative reward model baseline that does not incorporate data reweighting\footnote{\url{https://huggingface.co/BBQGOD/DeepSeek-GRM-16B}}.

\paragraph{Ablation baselines.}
Each ablation of \Cref{sec:ablation-source} keeps the backbone, the initialization, the evaluation protocol and the candidates of the proposed setting and changes one component. \emph{No reweighting} trains the judge with GRPO on the training set alone, with every sample weighted equally. \emph{Meta set only} trains it with GRPO on the meta set alone. \emph{Pooled training and meta sets} pools the two sets into one training set and trains with GRPO without reweighting. \emph{Strong-policy SFT} uses the data of the strong policy models, o4-mini (high) and DeepSeek V3.2 Speciale, which supply the candidates of the meta set, as training data instead of as the meta set: it fine-tunes the judge by supervised fine-tuning on all 442 judging pairs built from their candidates, with the correct selection as the target. \emph{Domain filter} discards training samples instead of weighting them: within each of the six domains it keeps a random subset so that the kept samples have the domain composition of the meta set, drops the rest, and trains with GRPO on the kept 42.7\% of the training set without reweighting. \emph{Inverse-frequency heuristic} fixes the weight of each of the six domains in proportion to $1/(\text{domain size})$, normalized to sum to one, in place of the learned domain weights. \emph{BigCodeBench meta set} builds the meta set from BigCodeBench problems and candidates instead of LiveCodeBench ones, while the training set and the evaluation stay on LiveCodeBench. \emph{Direct selection} runs the judge without chain-of-thought reasoning at inference, so that it emits its selection directly. \emph{Untrained Qwen3-1.7B} and \emph{SFT warmup} are the judge before any training and after the supervised warmup that initializes every GRPO run.

\paragraph{Training.}
The proposed bi-level optimization-based data-reweighted training framework is implemented using the Betty library~\citep{choe2023betty}. Dataset preparation and evaluation follow the official repositories for LiveCodeBench~\citep{jain2025livecodebench}\footnote{\url{https://github.com/LiveCodeBench/LiveCodeBench}} and BigCodeBench~\citep{zhuo2024bigcodebench}\footnote{\url{https://github.com/bigcode-project/bigcodebench}}.
We use Adam~\citep{Kingma2014AdamAM} as the optimizer, with a learning rate of $10^{-6}$ for the lower-level optimization and a meta-learning rate of $10^{-4}$ for the upper-level optimization. Weight decay is set to $0$, and the batch size is $1$. Gradient accumulation is applied with 16 steps for lower-level optimization and 8 steps for upper-level optimization.
All models are trained for 20{,}000 iterations. Preference optimization experiments are conducted using a single NVIDIA A100 GPU, while reinforcement learning experiments use two NVIDIA A100 GPUs, with one GPU dedicated to rollout and the other to training. All experiments are performed using bfloat16 (bf16) precision.
The rollout is conducted with vLLM~\citep{kwon2023efficient} as the efficient inference engine.
For candidate generation, we sample four candidates per problem for o4-mini (high) and DeepSeek V3.2 Speciale, and eight candidates per problem for Qwen2.5-Coder-32B and Qwen3-Coder-30B-A3B. All candidates are decoded with temperature $0.7$, top-$p$ $0.8$, top-$k$ $20$, repetition penalty $1.05$, and a maximum output length of $8{,}192$ tokens. Candidate selection is performed using the pairwise voting procedure described in \Cref{sec:method} with $R{=}8$ rounds.
We adopt LoRA~\citep{hu2022lora} for parameter-efficient fine-tuning, with rank 32, scaling factor 64, and dropout rate set to $0.0$. LoRA adapters are applied to the query, key, value, and output projection layers, as well as the gate, up, and down projection layers of the transformer.
For instance-level reweighting, the instance net is a multilayer perceptron (MLP) with two hidden layers of 10 units.

We consider several preference optimization approaches for training the judge. Direct Preference Optimization (DPO; \citealp{rafailov2023direct}) directly optimizes preferences without relying on an explicit reward model; in our experiments, we set the temperature parameter to $\beta_{\text{D}}=0.1$. Kahneman--Tversky Optimization (KTO; \citealp{ethayarajh2024kto}) is an alignment method that maximizes generation utility without requiring paired preference data. For KTO, we use $\beta_{\text{K}}=0.1$, and the class weights $\lambda_{\text{D}}$ and $\lambda_{\text{U}}$ of the chosen and rejected samples are balanced. Odds Ratio Preference Optimization (ORPO; \citealp{hong2024orpo}) is a reference-model-free, monolithic preference optimization approach that removes the need for a separate preference alignment stage. We set $\lambda_{\text{O}}=0.1$ for ORPO to balance the supervised fine-tuning loss and the preference loss.
In addition to preference optimization, we train the judge using Group Relative Policy Optimization (GRPO; \citealp{shao2024deepseekmath}), an online reinforcement learning method with verifiable rewards (RLVR). GRPO optimizes policies using relative comparisons within groups, and we set the group size to 16 in all experiments.

\section{More Experimental Results}
\label{app:more-results}

\paragraph{Cost and accuracy of frontier judges.}
\label{app:frontier-table}
\Cref{tab:frontier-judges} gives the exact accuracy and cost of the judges that \Cref{sec:frontier-judges} compares in \Cref{fig:cost-accuracy}.

\begin{table*}[!htb]
  \centering
  \scriptsize
  \caption{Cost and accuracy of frontier LLM judges on LiveCodeBench. Last column: cost relative to \our{}.}
  \label{tab:frontier-judges}
  \begin{tabularx}{0.75\linewidth}{l<{\hspace{8pt}}lYYYY}
      \toprule
      Judge & Protocol & Overall & Hard & Cost (USD) & vs.\ \our{} \\
      \midrule
      Random            & Uniform selection   & 76.0          & 54.9          & \$0.00 & -- \\
      Claude Haiku 4.5  & Listwise           & 83.2          & 70.5          & \$10.97 & 48$\times$ \\
      Claude Sonnet 4.6 & Listwise           & 84.7          & 72.1          & \$31.19 & 136$\times$ \\
      Claude Haiku 4.5  & Pairwise ($R{=}8$) & 85.5          & 72.1          & \$62.85 & 273$\times$ \\
      Claude Sonnet 4.6 & Pairwise ($R{=}8$) & \textbf{86.3} & \textbf{73.8} & \$106.48 & 463$\times$ \\
      \rowcolor{orange!12}\our{} 1.7B (ours) & Pairwise ($R{=}8$) & 84.7 & \textbf{73.8} & \textbf{\$0.23} & \textbf{1$\times$} \\
      \bottomrule
    \end{tabularx}
\end{table*}

\paragraph{Per-difficulty breakdowns.}
\label{appendix:per-difficulty-results}
\label{appendix:per-split-policy}

The main text reports only Overall pass@1 in \Cref{tab:results,tab:results-policy} for compactness. This section provides the corresponding results broken down by Easy, Medium, and Hard problems.
\Cref{tab:results-full} expands the main leaderboard (\Cref{tab:results}) into Easy / Medium / Hard splits for every zero-shot and test-time scaling baseline, on both LiveCodeBench and BigCodeBench. \Cref{tab:results-policy-full} expands the cross-policy results (\Cref{tab:results-policy}) into the same Easy / Medium / Hard splits across all four policy models, additionally including the Random selection baseline. \Cref{tab:ablation-full} gives the same splits for the preference-optimization ablation (\Cref{tab:ablation}).

\begin{table*}[!t]
  \centering
  \scriptsize
  \caption{Full per-method, per-difficulty performance on LiveCodeBench and BigCodeBench.}
  \label{tab:results-full}
  \begin{tabularx}{\linewidth}{lYYYYlY}
  \toprule
  \multicolumn{5}{c}{LiveCodeBench} & \multicolumn{2}{c}{BigCodeBench} \\
  \cmidrule(lr){1-5} \cmidrule(lr){6-7}
  Method                & Overall       & Easy           & Medium         & Hard          & Method               & Overall       \\
  \midrule
  \rowcolor{gray!20}\multicolumn{7}{c}{\textit{Zero-shot Methods}} \\
  o4-mini (high)        & \textbf{77.1} & \textbf{100.0} & \textbf{89.7}  & \textbf{57.4} & o3-mini (medium)     & \textbf{33.1} \\
  Gemini-2.5-Pro-06-05  & 72.5          & \textbf{100.0} & 82.1           & 52.5          & o1 (high)            & 32.4          \\
  o4-mini (medium)      & 71.8          & 96.8           & 79.5           & 54.1          & o3-mini (high)       & 32.4          \\
  o3 (high)             & 71.8          & \textbf{100.0} & 71.8           & \textbf{57.4} & Claude-3.7-Sonnet    & 32.4          \\
  \midrule
  \rowcolor{gray!20}\multicolumn{7}{c}{\textit{Test-time Scaling Methods}} \\
  Skywork-o1 PRM        & 68.7          & \textbf{100.0} & 84.6           & 42.6          & Skywork-o1 PRM       & 35.2          \\
  DeepSeek GRM          & 83.2          & \textbf{100.0} & \textbf{94.9}  & 67.2          & DeepSeek GRM         & 33.8          \\
  \rowcolor{orange!12}\our{}                & \textbf{84.7} & \textbf{100.0} & 89.7           & \textbf{73.8} & \our{}               & \textbf{35.9} \\
  \bottomrule
  \end{tabularx}
\end{table*}

\begin{table}[!t]
  \centering
  \scriptsize
  \caption{Results on LiveCodeBench by policy model.}
  \label{tab:results-policy-full}
  \begin{tabularx}{\linewidth}{lYYYY}
  \toprule
  Method & Overall & Easy & Medium & Hard \\
  \midrule
  \rowcolor{gray!20}\multicolumn{5}{c}{\textit{o4-mini (high)}} \\
  Random & 75.8 & 98.4 & 80.8 & 61.1 \\
  Skywork-o1 PRM 7B & 74.1 & 96.8 & 82.1 & 57.4 \\
  DeepSeek GRM 16B & 75.6 & \textbf{100.0} & 79.5 & 60.7 \\
  \rowcolor{orange!12}\our{} 1.7B & \textbf{81.7} & \textbf{100.0} & \textbf{89.7} & \textbf{67.2} \\
  \midrule
  \rowcolor{gray!20}\multicolumn{5}{c}{\textit{DeepSeek V3.2 Speciale}} \\
  Random & 76.0 & \textbf{100.0} & 89.7 & 54.9 \\
  Skywork-o1 PRM 7B & 68.7 & \textbf{100.0} & 84.6 & 42.6 \\
  DeepSeek GRM 16B & 83.2 & \textbf{100.0} & \textbf{94.9} & 67.2 \\
  \rowcolor{orange!12}\our{} 1.7B & \textbf{84.7} & \textbf{100.0} & 89.7 & \textbf{73.8} \\
  \midrule
  \rowcolor{gray!20}\multicolumn{5}{c}{\textit{Qwen2.5-Coder-32B}} \\
  Random & 23.3 & 75.4 & 14.4 & \textbf{2.5} \\
  Skywork-o1 PRM 7B & 25.2 & 80.7 & 20.5 & 0.0 \\
  DeepSeek GRM 16B & 25.2 & 77.4 & \textbf{23.1} & 0.0 \\
  \rowcolor{orange!12}\our{} 1.7B & \textbf{26.7} & \textbf{83.9} & 20.5 & 1.6 \\
  \midrule
  \rowcolor{gray!20}\multicolumn{5}{c}{\textit{Qwen3-Coder-30B-A3B}} \\
  Random & 36.9 & 90.7 & 34.0 & 11.5 \\
  Skywork-o1 PRM 7B & 36.6 & 90.3 & 33.3 & 11.5 \\
  DeepSeek GRM 16B & 38.2 & 90.3 & \textbf{35.9} & \textbf{13.1} \\
  \rowcolor{orange!12}\our{} 1.7B & \textbf{38.9} & \textbf{93.5} & \textbf{35.9} & \textbf{13.1} \\
  \midrule
  \rowcolor{gray!20}\multicolumn{5}{c}{\textit{Average over all policy models}} \\
  Random & 53.0 & 91.1 & 54.7 & 32.5 \\
  Skywork-o1 PRM 7B & 51.2 & 92.0 & 55.1 & 27.9 \\
  DeepSeek GRM 16B & 55.6 & 91.9 & 58.4 & 35.3 \\
  \rowcolor{orange!12}\our{} 1.7B & \textbf{58.0} & \textbf{94.4} & \textbf{59.0} & \textbf{38.9} \\
  \bottomrule
  \end{tabularx}
\end{table}

\begin{table}[!htb]
  \centering
  \scriptsize
  \caption{Full per-difficulty results of the preference-optimization ablation on LiveCodeBench.}
  \label{tab:ablation-full}
  \begin{tabularx}{\linewidth}{lYYYY}
  \toprule
  Method & Overall & Easy & Medium & Hard \\
  \midrule
  \rowcolor{gray!20}\multicolumn{5}{c}{\textit{DPO}} \\
  No reweighting & 81.7 & \textbf{100.0} & 92.3 & 65.6 \\
  Domain reweighting & \textbf{83.2} & \textbf{100.0} & \textbf{94.9} & 67.2 \\
  Instance table & 82.4 & \textbf{100.0} & \textbf{94.9} & 65.6 \\
  Instance net & \textbf{83.2} & \textbf{100.0} & 92.3 & \textbf{68.9} \\
  \midrule
  \rowcolor{gray!20}\multicolumn{5}{c}{\textit{KTO}} \\
  No reweighting & 83.2 & \textbf{100.0} & \textbf{94.9} & 67.2 \\
  Domain reweighting & 83.2 & \textbf{100.0} & 92.3 & 68.9 \\
  Instance table & \textbf{84.0} & \textbf{100.0} & 92.3 & \textbf{70.5} \\
  Instance net & \textbf{84.0} & \textbf{100.0} & \textbf{94.9} & 68.9 \\
  \midrule
  \rowcolor{gray!20}\multicolumn{5}{c}{\textit{ORPO}} \\
  No reweighting & 81.7 & \textbf{100.0} & 92.3 & 65.6 \\
  Domain reweighting & 82.4 & \textbf{100.0} & 89.7 & 68.9 \\
  Instance table & \textbf{83.2} & \textbf{100.0} & 89.7 & \textbf{70.5} \\
  Instance net & \textbf{83.2} & \textbf{100.0} & \textbf{94.9} & 67.2  \\
  \midrule
  \rowcolor{gray!20}\multicolumn{5}{c}{\textit{Average over all preference optimization methods}} \\
  No reweighting & 82.2 & \textbf{100.0} & 93.2 & 66.1 \\
  Domain reweighting & 82.9 & \textbf{100.0} & 92.3 & 68.3 \\
  Instance table & 83.2 & \textbf{100.0} & 92.3 & \textbf{68.9} \\
  Instance net & \textbf{83.5} & \textbf{100.0} & \textbf{94.0} & 68.3 \\
  \bottomrule
  \end{tabularx}
\end{table}

\paragraph{Ten inference seeds.}
\label{app:ten-seeds}
\Cref{tab:grpo-ablation} compares domain reweighting with no reweighting under GRPO over ten inference seeds, with pairwise voting at $R{=}8$. Domain reweighting gains 1.37 points Overall and 2.62 points on Hard problems. With the two judges paired by inference seed, the Overall gain has $t(9){=}2.65$, $p{=}0.027$ and a 95\% CI of $[0.20, 2.55]$, and the Hard gain has $t(9){=}3.36$, $p{=}0.008$ and a 95\% CI of $[0.86, 4.39]$.

\begin{table}[t]
  \centering
  \scriptsize
  \caption{GRPO judges on LiveCodeBench over five inference seeds (mean $\pm$ standard deviation): the initialization and the instance-level reweighting strategies, with no reweighting and domain reweighting.}
  \label{tab:grpo-instance}
  \begin{tabular}{lcc}
    \toprule
    Method & Overall & Hard \\
    \midrule
    Untrained Qwen3-1.7B & 79.85 $\pm$ 1.28 & 61.97 $\pm$ 2.13 \\
    SFT warmup & 82.44 $\pm$ 1.53 & 67.21 $\pm$ 2.59 \\
    \midrule
    No reweighting & 83.21 $\pm$ 1.08 & 68.52 $\pm$ 1.80 \\
    Domain reweighting & \textbf{84.43 $\pm$ 0.87} & \textbf{71.48 $\pm$ 1.47} \\
    Instance table & 83.97 $\pm$ 0.54 & 69.51 $\pm$ 1.47 \\
    Instance net & 84.27 $\pm$ 0.68 & 69.83 $\pm$ 1.47 \\
    \bottomrule
  \end{tabular}
\end{table}

\paragraph{Instance-level reweighting under GRPO.}
\label{app:grpo-instance}
Under GRPO, the instance table and the instance net reach 83.97\% and 84.27\% Overall and 69.51\% and 69.83\% Hard, between no reweighting and domain reweighting (\Cref{tab:grpo-instance}).

\paragraph{Comparison with public reward models.}
\label{app:public-rms}

\Cref{tab:public-rms} compares \our{} with four public reward models on the same 131 LiveCodeBench problems and the same candidates from the four policy models of \Cref{tab:results-policy}; each number is the average over the four policy models. Skywork-Reward-V2~\citep{liu2025skyworkrewardv2} and AceCodeRM~\citep{zeng2025acecoder} score each candidate alone, and the highest-scored candidate is selected. RM-R1~\citep{chen2025rmr1} compares two candidates at a time, like \our{}: it runs the same pairwise voting with $R{=}8$, with the prompt from its model card, greedy decoding, and up to 8{,}192 new tokens, and the candidate with the most wins is selected; a judgment that ends without a verdict gives no win. When a reward model ties several candidates, its reported accuracy is the mean over the tied candidates. RM-R1 is averaged over five inference seeds, which set the order in which each pair is shown. Averaged over the four policy models, \our{} reaches the highest Overall and Hard accuracy of all judges in \Cref{tab:public-rms}, above four that are four to nine times larger.

\begin{figure*}[!t]
  \centering
  \includegraphics[width=0.88\textwidth]{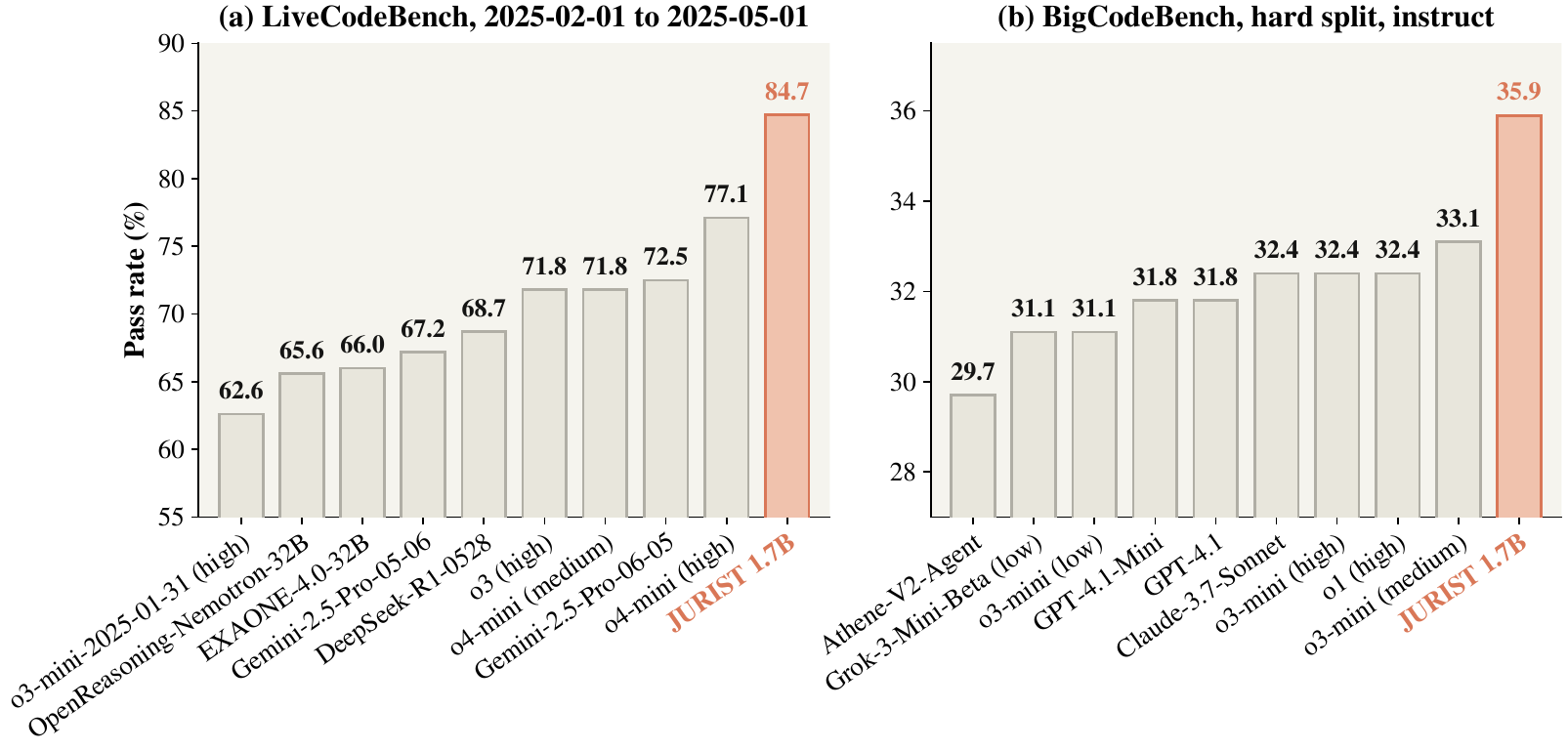}
  \caption{Public leaderboard results on LiveCodeBench and BigCodeBench.}
  \label{fig:leaderboard}
\end{figure*}

\paragraph{Comparison with public leaderboards.}
\label{app:leaderboard}

\Cref{fig:leaderboard} places \our{} among the methods on the public LiveCodeBench and BigCodeBench leaderboards. The public leaderboard ranks all submitted methods, single-shot and test-time scaling alike, by overall accuracy on a fixed problem set.

\begin{figure}[t]
  \centering
  \includegraphics[width=\linewidth]{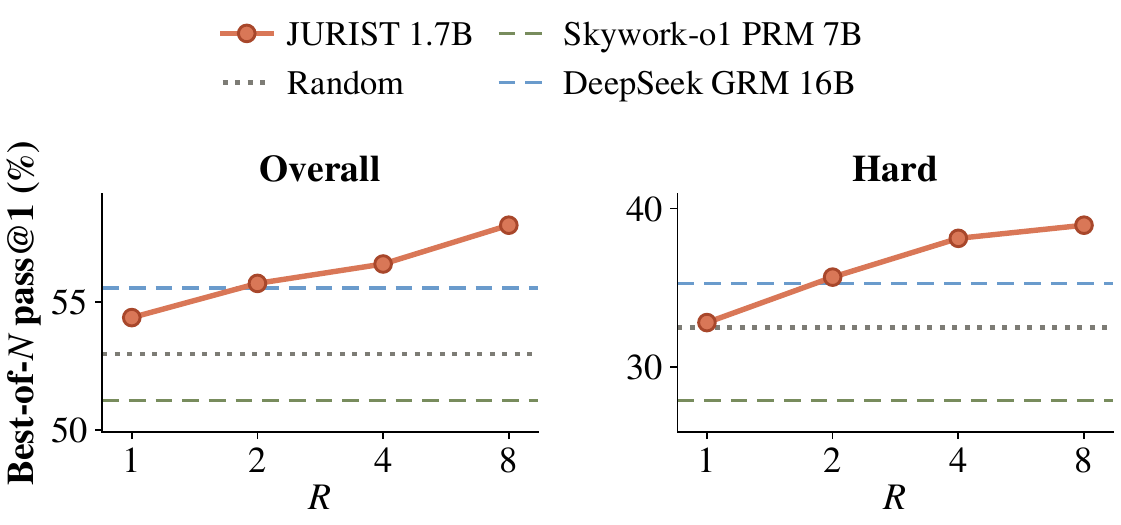}
  \caption{Voting-round scaling on LiveCodeBench, averaged over four policy models.}
  \label{fig:r-sweep}
\end{figure}

\paragraph{Voting rounds.}
\label{app:voting-rounds}
In \Cref{fig:r-sweep}, the four-policy accuracy of \our{} grows monotonically with $R$ and exceeds DeepSeek GRM at $R{=}2$ on both Overall and Hard, so $R$ can be matched to the compute budget without retraining.

\begin{figure*}[!t]
  \centering
  \includegraphics[width=\linewidth]{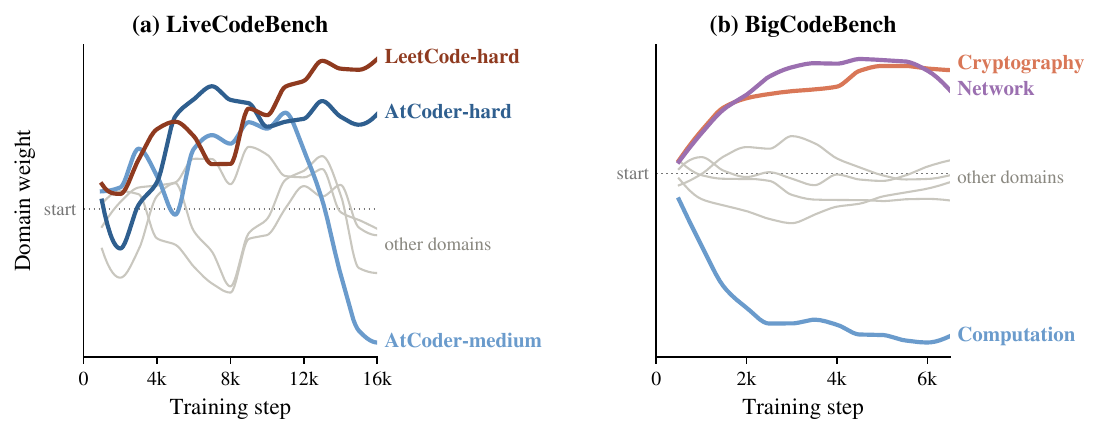}
  \caption{Learned domain weights during training on (a) LiveCodeBench and (b) BigCodeBench.}
  \label{fig:domain-weights}

  \vspace{10pt}
  \includegraphics[width=\textwidth]{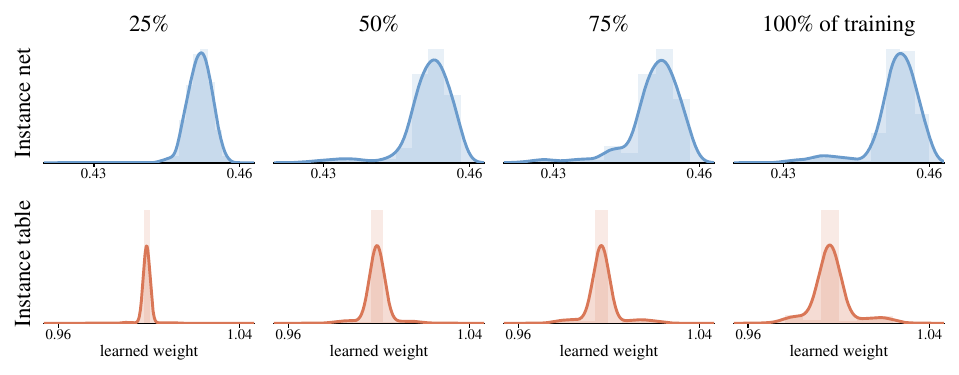}
  \caption{Distribution of the learned instance-level data weights over the training samples at 25\%, 50\%, 75\% and 100\% of training, for the KTO-trained Qwen2.5-Coder-14B judge on LiveCodeBench: a smoothed density over the ten-bin histogram (faint bars); x-axis: weight.}
  \label{fig:instance-weights}
\end{figure*}

\paragraph{Evolution of data weights.}
\label{appendix:evolution-of-data-weights}
The learned weights provide a complementary view of the behavior summarized in \Cref{app:operationalization}. Domain weights start uniformly and diverge during training, emphasizing informative and challenging domains such as hard AtCoder tasks while downweighting easier or less relevant ones (\Cref{fig:domain-weights}). At the instance level, both the instance net and instance table exhibit increasing weight dispersion over time (\Cref{fig:instance-weights}), with training focus shifting from lower-quality to more informative samples.

\section{Gradient Analysis}
\label{app:operationalization}

This appendix derives the upper-level update of \Cref{sec:method} step by step and shows how its two factors address the three shifts. We write $\alpha_i$ for $\alpha_{(X_i,Y_i)}$ and $\mathcal{L}_i(\phi)$ for $\mathcal{L}(X_i,Y_i;\phi)$.

\paragraph{One-step update.}
\emph{Step 1: lower-level step.} The weighted training loss is $\mathcal{L}_{\text{tr}}(\mathcal{D}_{\text{tr}},\phi,\alpha) = \sum_j \alpha_j\,\mathcal{L}_j(\phi)$, so its gradient at $\phi^{(t)}$ is $\sum_j \alpha_j\, g_j$ with $g_j = \nabla_\phi\mathcal{L}_j(\phi^{(t)})$ as in \Cref{sec:method}. One optimizer step with learning rate $\beta_1$ and preconditioner $P$ gives
\begin{equation}
\label{eq:app-step}
\phi^{(t+1)}(\alpha) = \phi^{(t)} - \beta_1 P \sum_j \alpha_j\, g_j .
\end{equation}
For SGD, $P = I$; for Adam,
\begin{equation*}
P = \operatorname{diag}\big(1/(\sqrt{\hat v}+\epsilon)\big)
\end{equation*}
with the bias-corrected second-moment estimate $\hat v$. Since $\hat v$ accumulates past gradients, $P$ depends on $\alpha$; we treat it as a constant within the step, a first-order approximation, and ignore momentum.

\emph{Step 2: derivative of the step.} In \Cref{eq:app-step}, $\phi^{(t)}$, $P$ and every $g_j$ are computed before the step and do not depend on $\alpha$, and only the term $j=i$ contains $\alpha_i$. Hence
\begin{equation*}
\frac{\partial \phi^{(t+1)}}{\partial \alpha_i} = -\beta_1 P\, g_i .
\end{equation*}

\emph{Step 3: chain rule.} The meta loss depends on $\alpha_i$ only through $\phi^{(t+1)}$. With the meta-loss gradient $g_{\text{meta}}$ at $\phi^{(t+1)}$, as defined in \Cref{sec:method},
\begin{align*}
\frac{\partial \mathcal{L}_{\text{meta}}}{\partial \alpha_i}
&= \Big\langle \frac{\partial \phi^{(t+1)}}{\partial \alpha_i},\, g_{\text{meta}} \Big\rangle \\
&= \big\langle -\beta_1 P g_i,\, g_{\text{meta}} \big\rangle
= -\beta_1 \big\langle g_i,\, P g_{\text{meta}} \big\rangle ,
\end{align*}
where the last equality uses $P^\top = P$, since $P$ is diagonal. Negating both sides gives \Cref{eq:hypergrad}.

\emph{Step 4: weight update.} Substituting into the step $\alpha_i^{(t+1)} = \alpha_i^{(t)} - \beta_2\,\partial\mathcal{L}_{\text{meta}}/\partial\alpha_i$ gives
\begin{equation*}
\alpha_i^{(t+1)} = \alpha_i^{(t)} + \beta_1\beta_2 \big\langle g_i,\, P g_{\text{meta}} \big\rangle .
\end{equation*}
The weight of sample $i$ rises exactly when the inner product is positive. For small $\beta_1$, $g_{\text{meta}}$ at $\phi^{(t+1)}$ is close to its value at $\phi^{(t)}$, so to first order in $\beta_1$ the weight rises when the step $-P g_i$ that lowers the loss of sample $i$ also lowers the meta loss. This is the look-ahead of learning to reweight examples~\citep{ren2018learning}; Betty~\citep{choe2023betty} evaluates the product with a finite difference.

\paragraph{Margin form.}
\emph{Step 1: margin.} Suppose the per-sample loss depends on $\phi$ only through a scalar margin, $\mathcal{L}_i(\phi) = f(m_i(\phi))$, with $f$ decreasing. Define the learning signal $\psi(m) = -f'(m) \ge 0$ and $\psi_i = \psi(m_i(\phi^{(t)}))$.

\emph{Step 2: gradient.} By the chain rule, $g_i = f'(m_i)\,\nabla_\phi m_i = -\psi_i\,\nabla_\phi m_i$.

\emph{Step 3: substitution.} Substituting $g_i$ into \Cref{eq:hypergrad} and moving $\psi_i$ out of the inner product and the minus sign into its second argument,
\begin{align*}
-\frac{\partial \mathcal{L}_{\text{meta}}}{\partial \alpha_i}
&= \beta_1 \big\langle -\psi_i \nabla_\phi m_i,\, P g_{\text{meta}} \big\rangle \\
&= \beta_1\, \psi_i\, \big\langle \nabla_\phi m_i,\, -P g_{\text{meta}} \big\rangle ,
\end{align*}
which is \Cref{eq:factor}. The vector $-P g_{\text{meta}}$ is the preconditioned direction that lowers the meta loss, so the alignment is positive when raising the margin of sample $i$ moves $\phi$ along that direction.

\paragraph{Learning signal of each objective.}
Each case follows from $\psi = -f'(m)$. Let $q = \sigma(m)$, and recall $\sigma'(m) = q(1-q)$.
\begin{itemize}[topsep=2pt, itemsep=3pt, parsep=0pt, leftmargin=*]
\item \textbf{DPO.} With chosen and rejected judgments $y^+, y^-$, reference policy $\pi_{\text{ref}}$ and temperature $\beta_{\text{D}}$, the margin and loss are
\begin{gather*}
m = \beta_{\text{D}}\Big[\log\frac{\pi_\phi(y^+)}{\pi_{\text{ref}}(y^+)} - \log\frac{\pi_\phi(y^-)}{\pi_{\text{ref}}(y^-)}\Big], \\
f(m) = -\log\sigma(m), \\
f'(m) = -(1-q), \quad \psi = 1-q .
\end{gather*}
It vanishes only as $q \to 1$ and peaks as $q \to 0$.
\item \textbf{KTO.} With implicit reward $r = \log\frac{\pi_\phi(y)}{\pi_{\text{ref}}(y)}$ and a detached reference point $z$, a desirable and an undesirable judgment have the losses
\begin{gather*}
\mathcal{L} = \lambda_{\text{D}}\big(1-\sigma(\beta_{\text{K}}(r-z))\big), \\
\mathcal{L} = \lambda_{\text{U}}\big(1-\sigma(\beta_{\text{K}}(z-r))\big).
\end{gather*}
With $m = \beta_{\text{K}}(r-z)$ or $m = \beta_{\text{K}}(z-r)$, and $\lambda = \lambda_{\text{D}}$ or $\lambda_{\text{U}}$,
\begin{gather*}
f(m) = \lambda\big(1-\sigma(m)\big), \\
f'(m) = -\lambda\, q(1-q), \quad \psi = \lambda\, q(1-q) .
\end{gather*}
The margin is that of one judgment, and the signal vanishes at both ends.
\item \textbf{ORPO.} ORPO~\citep{hong2024orpo} uses the length-normalized likelihood and the odds
\begin{gather*}
\bar P_\phi(y) = \exp\Big(\frac{1}{|y|}\sum_t \log\pi_\phi(y_t \mid y_{<t})\Big), \\
\mathrm{odds}_\phi(y) = \frac{\bar P_\phi(y)}{1-\bar P_\phi(y)} .
\end{gather*}
With the margin and $q$
\begin{gather*}
m_{\text{OR}} = \log\mathrm{odds}_\phi(y^+) - \log\mathrm{odds}_\phi(y^-), \\
q = \sigma(m_{\text{OR}}),
\end{gather*}
the loss and its gradient are
\begin{align*}
\mathcal{L} &= -\log\bar P_\phi(y^+) - \lambda_{\text{O}}\log\sigma(m_{\text{OR}}), \\
\nabla_\phi\mathcal{L} &= -\nabla_\phi\log\bar P_\phi(y^+) - \lambda_{\text{O}}(1-q)\nabla_\phi m_{\text{OR}} .
\end{align*}
The margin term follows \Cref{eq:factor} with $\psi = \lambda_{\text{O}}(1-q)$. The likelihood term adds $\beta_1\langle \nabla_\phi\log\bar P_\phi(y_i^+),\, -P g_{\text{meta}}\rangle$, an alignment term with no learning signal.
\item \textbf{GRPO.} GRPO samples a group of judgments per sample and normalizes the rewards within the group. The next paragraph shows that, in expectation, it is a margin loss with $\psi = \sqrt{p(1-p)}$.
\end{itemize}

\paragraph{GRPO as a margin loss.}
Fix one training sample $i$, and let $p = p_i$ be the probability that a judgment selects a passing candidate, with $0 < p < 1$. Under \Cref{eq:reward}, a judgment receives $r_+ = 1$ for a passing selection, $r_- = 0.5$ for a failing one, and $0$ for a format error; assume no format error, so the reward takes two values. For two-valued rewards and a large group, \citet{mroueh2025reinforcement} shows that GRPO without clipping maximizes (their Eq.~5)
\begin{gather*}
\omega_+(p)\,p_\phi - \omega_-(p)\,(1-p_\phi), \\
\omega_+(p) + \omega_-(p) = 1/\sqrt{p(1-p)},
\end{gather*}
where $p_\phi$ is the success probability under $\pi_\phi$. Normalizing the rewards removes their shift and scale, so the values $1$ and $0.5$ give the same weights. At $p_\phi = p$, its gradient is $\nabla_\phi p_\phi/\sqrt{p(1-p)}$. With the margin $m_i = \log\frac{p_\phi}{1-p_\phi}$, $\nabla_\phi p_\phi = p_\phi(1-p_\phi)\,\nabla_\phi m_i$, so GRPO has the margin form with
\begin{equation*}
\psi_i = \sqrt{p_i(1-p_i)} .
\end{equation*}
We ignore the KL term, per-token averaging, and a constant advantage scale, which scales every $\psi_i$ equally. Two cases fall outside this analysis:
\begin{itemize}[topsep=2pt, itemsep=2pt, parsep=0pt, leftmargin=*]
\item a group with a format error, whose reward takes a third value;
\item a finite group whose judgments all receive the same reward, so $\mathrm{std}(r) = 0$. Our implementation gives such a group a constant advantage of $\pm 0.1$, so it has a small learning signal outside the margin form.
\end{itemize}

\paragraph{Domain weights.}
\emph{Step 1: domain gradient.} Every sample of domain $k$ has the weight $\alpha_k$. Our implementation averages the weighted losses over the minibatch, and we absorb the constant $1/B$ for a batch of $B$ samples into $\beta_1$; the sums over $i \in k$ below run over the samples of domain $k$ in the batch. The lower-level step is $\phi^{(t+1)} = \phi^{(t)} - \beta_1 P \sum_k \alpha_k \sum_{i\in k} g_i$. Steps 2 and 3 of the one-step update, with $\sum_{i\in k} g_i$ in place of $g_i$, and the margin form give
\begin{gather*}
-\frac{\partial \mathcal{L}_{\text{meta}}}{\partial \alpha_k} = \beta_1 a_k, \\
a_k = \sum_{i\in k}\psi_i\big\langle\nabla_\phi m_i,\, -P g_{\text{meta}}\big\rangle .
\end{gather*}
\emph{Step 2: softmax.} With $\alpha = \operatorname{softmax}(u)$, $\partial\alpha_l/\partial u_k = \alpha_l(\delta_{lk}-\alpha_k)$. By the chain rule,
\begin{align*}
-\frac{\partial \mathcal{L}_{\text{meta}}}{\partial u_k}
&= \sum_l \beta_1 a_l\, \alpha_l(\delta_{lk}-\alpha_k) \\
&= \beta_1\,\alpha_k\Big(a_k - \sum_l\alpha_l a_l\Big) .
\end{align*}
A domain therefore gains share when its summed signal-times-alignment $a_k$ exceeds the average $\sum_l\alpha_l a_l$ of the current mixture. Since $a_k$ is a sum, it also grows with the number of samples of domain $k$ in the batch.

\paragraph{Instance weights.}
\label{app:instance-weights}
\emph{Instance table.} Each sample has its own weight $\alpha_i$, so \Cref{eq:factor} is its update: the weight of a sample rises with the product of its learning signal and its alignment. Only the samples drawn in the batch are updated at a step.

\emph{Instance net.} The net maps the loss of a sample, $\mathcal{L}_i = \mathcal{L}(X_i, Y_i; \phi^{(t)})$, which is fixed during the step, to its weight $\alpha_i = \mathrm{MLP}_\theta(\mathcal{L}_i)$, so the meta loss depends on $\theta$ through the weights of the batch. By the chain rule through \Cref{eq:factor},
\begin{equation*}
-\frac{\partial \mathcal{L}_{\text{meta}}}{\partial \theta}
= \beta_1 \sum_{i} \psi_i \big\langle \nabla_\phi m_i, -P g_{\text{meta}} \big\rangle \, \nabla_\theta \alpha_i .
\end{equation*}
Each sample's signal-times-alignment therefore moves the net's output at that sample's loss, and samples with similar losses pool their updates: the net learns, as a function of the loss, the signal-times-alignment of the samples at that loss. The learning signal and the loss both depend on the judge through the sample's margin, so the net can express the learning signal; the alignment varies among samples with the same loss, and the net expresses only the part of it that correlates with the loss (\Cref{tab:factor-map}).

\begin{figure}[!t]
  \centering
  \includegraphics[width=\linewidth]{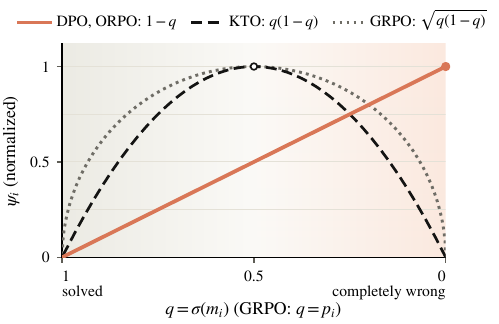}
  \caption{Learning signal $\psi_i$ of each objective (\Cref{tab:signal}), normalized to its maximum, from samples the judge solves ($q = 1$) to samples it gets completely wrong ($q = 0$). For DPO, KTO and ORPO, $q = \sigma(m_i)$; for GRPO, $q = p_i$ in the large-group limit.}
  \label{fig:psi-curves}
\end{figure}

\paragraph{The three shifts.}
Each shift is addressed mainly by one factor of \Cref{eq:factor}:
\begin{itemize}[topsep=2pt, itemsep=3pt, parsep=0pt, leftmargin=*]
\item \textbf{Difficulty mismatch:} mainly the learning signal, which is small on the easy samples the judge has settled; the alignment helps, since the meta set holds harder problems and medium and hard problems align alike with it.
\item \textbf{Task mismatch:} the alignment, since the meta set holds problems released closer to the test period.
\item \textbf{Trajectory mismatch:} the alignment, since the meta set holds candidates from stronger models.
\end{itemize}
The two mismatches share one inner product, and the derivation does not tell their contributions apart. The instance table can express both factors but updates a weight only when its sample is drawn, and the instance net sees the alignment only through its correlation with the loss. How the learning signal treats the two ends of $q$ depends on the objective (\Cref{fig:psi-curves}): every objective gives a sample the judge has solved a small signal; DPO and ORPO give the largest signal to a sample the judge gets completely wrong, while KTO and GRPO concentrate the signal in between. In terms of the judge's accuracy $p_i$ on a sample, which is $q$ for GRPO, easy problems lie near $p_i = 1$ and most hard problems in between: empirically, the judge solves most easy problems in every judgment, and most hard problems receive both passing and failing selections (\Cref{tab:exp1-psi}).

\paragraph{Empirical analysis.}
We now measure the three shifts in the data (\Cref{tab:shift-composition}), the errors of the GRPO judge without reweighting over five inference seeds, and both factors of \Cref{eq:factor}.

\begin{table}[!t]
  \centering
  \scriptsize
  \caption{Distribution shifts on LiveCodeBench. Test set: 31 easy, 39 medium and 61 hard problems.}
  \label{tab:shift-composition}
  \begin{tabular}{lccc}
    \toprule
    Shift & Training & Meta & Test \\
    \midrule
    Hard problems & 23.0\% & 27.9\% & 46.6\% \\
    AtCoder-hard & 9.3\% & 21.3\% & 34.4\% \\
    Trajectory source & Qwen2.5/3-Coder & o4-mini, DeepSeek & DeepSeek \\
    \bottomrule
  \end{tabular}
\end{table}

The training set contains 23.0\% Hard problems, compared with 46.6\% at test, while the meta set lies between them at 27.9\%. AtCoder-hard problems show the same progression, increasing from 9.3\% of the training set to 21.3\% of the meta set and 34.4\% of the test set. The trajectory sources shift from Qwen2.5/3-Coder in the training set toward o4-mini and DeepSeek in the meta set and DeepSeek at test time. These composition gaps are reflected in the judge's errors: the no-reweighting judge's error rate reaches 31.5\% on Hard problems, and Hard problems account for 86.5\% of its errors despite comprising 46.6\% of the test set. Finally, the learned domain weights move toward the same under-represented region (\Cref{fig:domain-weights}).

\begin{figure*}[!t]
  \centering
  \includegraphics[width=\textwidth]{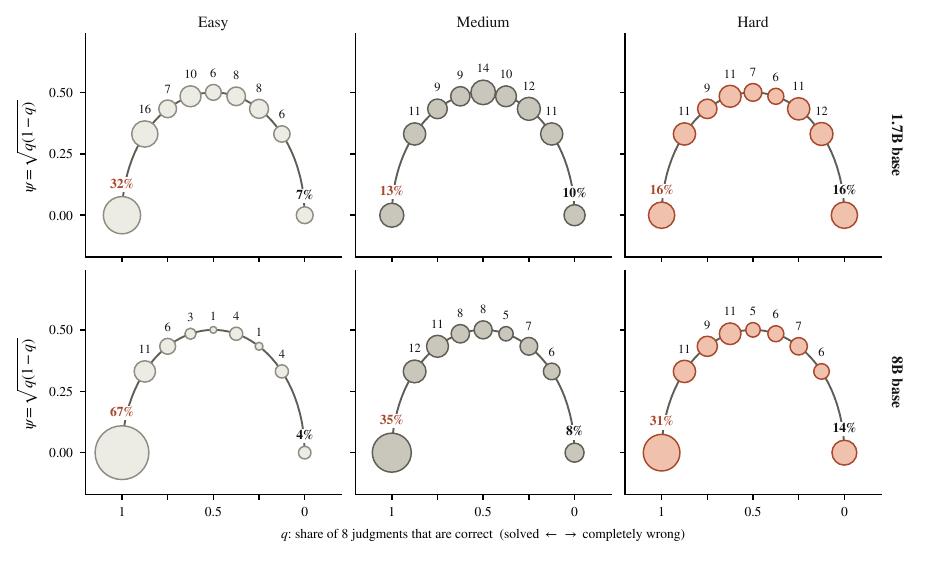}
  \caption{Share of training samples at each value of $q$, the share of 8 judgments that are correct, placed on the learning-signal curve $\psi = \sqrt{q(1-q)}$ of GRPO, for the 1.7B and 8B base judges. Bubble area and number: percentage of the samples of that difficulty. Bold: samples at $q = 1$ and $q = 0$, which carry no signal.}
  \label{fig:psi-bubbles}
\end{figure*}

\begin{figure}[!t]
  \centering
  \includegraphics[width=\linewidth]{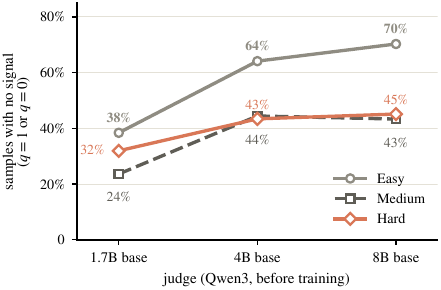}
  \caption{Share of training samples with no learning signal ($q = 1$ or $0$: all 8 judgments correct or all wrong) for Qwen3 base judges of three sizes before training.}
  \label{fig:nosignal-scale}
\end{figure}

\paragraph{Measured learning signal.}
We judge each training sample 8 times, take $p_i$ as the share of passing selections, and compute $\psi_i = \sqrt{p_i(1-p_i)}$ (\Cref{tab:exp1-psi}). Easy problems carry the smallest signal for both judges, because the judge solves most of them in every judgment: the 8B base judge selects a passing candidate in all 8 judgments on 66.7\% of easy problems, against 30.9\% of hard ones. The same holds where GRPO starts: after SFT warmup, the 1.7B judge has its smallest signal on easy problems (0.241, against 0.338 medium and 0.275 hard), and 41.7\% of easy problems carry none. Medium and hard problems carry similar signal for both judges (0.326 and 0.284 for 1.7B, 0.238 and 0.232 for 8B). By platform, $\psi$ rises from easy to medium to hard on LeetCode (0.151, 0.265, 0.296); on AtCoder it peaks at medium (0.211) and falls on hard (0.146), whose $p = 0.713$ is as high as that of medium (0.723). Most hard problems lie strictly between the two ends of \Cref{fig:psi-curves}: on LeetCode, 69.7\% of hard problems receive both passing and failing selections, against 38.8\% of easy problems. \Cref{fig:psi-bubbles} shows the per-sample distribution behind \Cref{tab:exp1-psi}: for the 8B base judge, two thirds of the easy problems sit at the zero-signal end $q = 1$. Saturation grows with judge strength (\Cref{fig:nosignal-scale}): the share of easy problems with no signal rises from 38\% for the 1.7B base judge to 64\% for 4B and 70\% for 8B, while medium and hard problems level off at 43--45\% from 4B on.

\begin{table}[!t]
  \centering
  \scriptsize
  \caption{Mean probability $p$ that the judge selects a passing candidate over 8 judgments, and mean learning signal $\psi = \sqrt{p(1-p)}$, on the training samples, for the 1.7B and 8B base judges before training and for the 1.7B judge after SFT warmup, where GRPO starts (120 samples per difficulty). Bold: largest $\psi$ per judge.}
  \label{tab:exp1-psi}
  \setlength{\tabcolsep}{5pt}
  \begin{tabularx}{\linewidth}{p{1.25cm}p{2.3cm}YY}
    \toprule
    Judge & Difficulty / domain & $p$ & $\psi$ \\
    \midrule
    \rowcolor{gray!20}\multicolumn{4}{c}{\textit{By difficulty}} \\
    1.7B base & Easy   & 0.663 & 0.257 \\
              & Medium & 0.505 & \textbf{0.326} \\
              & Hard   & 0.497 & 0.284 \\
    \addlinespace[3pt]
    8B base   & Easy   & 0.849 & 0.118 \\
              & Medium & 0.668 & \textbf{0.238} \\
              & Hard   & 0.618 & 0.232 \\
    \addlinespace[3pt]
    1.7B SFT  & Easy   & 0.590 & 0.241 \\
              & Medium & 0.548 & \textbf{0.338} \\
              & Hard   & 0.609 & 0.275 \\
    \midrule
    \rowcolor{gray!20}\multicolumn{4}{c}{\textit{By domain}} \\
    8B base   & LeetCode Easy   & 0.820 & 0.151 \\
              & LeetCode Medium & 0.613 & 0.265 \\
              & LeetCode Hard   & 0.546 & \textbf{0.296} \\
    \addlinespace[3pt]
              & AtCoder Easy    & 0.865 & 0.100 \\
              & AtCoder Medium  & 0.723 & 0.211 \\
              & AtCoder Hard    & 0.713 & 0.146 \\
    \bottomrule
  \end{tabularx}
\end{table}

\begin{figure*}[!t]
  \centering
  \includegraphics[width=\textwidth]{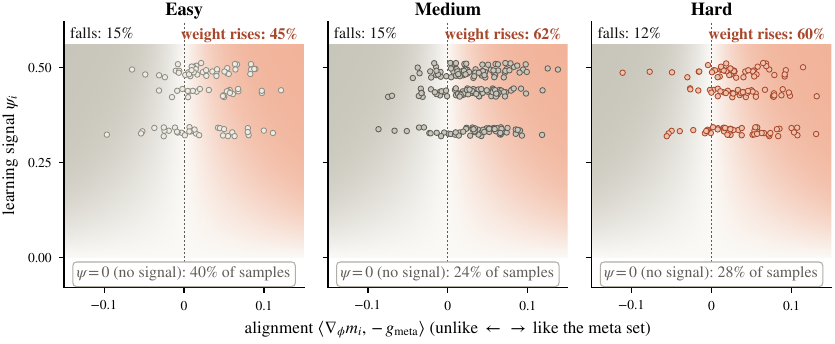}
  \caption{Training samples of each difficulty on the update map of \Cref{fig:update-map}, for the judge midway through training with domain reweighting. Each point is a sample with $\psi_i > 0$, jittered vertically. Top: share of samples whose weight falls or rises; bottom: share with $\psi_i = 0$. The three shares of a panel sum to 100\%.}
  \label{fig:update-samples}
\end{figure*}

\begin{figure*}[!t]
  \centering
  \includegraphics[width=0.9\textwidth]{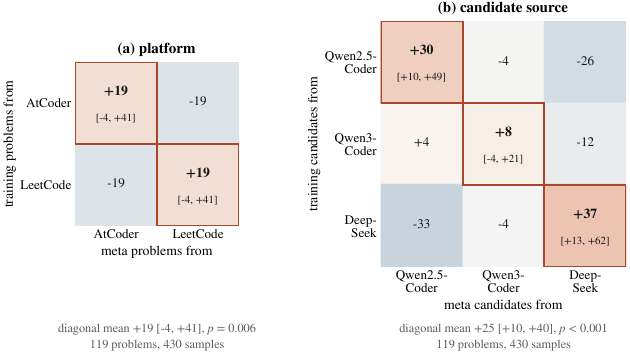}
  \caption{Alignment of training samples with meta samples of each group, after the common direction of the meta gradient is removed, double-centered ($\times 10^{-3}$, $P = I$). (a) Platform and (b) candidate source, for the judge midway through training with domain reweighting. Diagonal: same-group excess with a 95\% interval from resampling problems; $p$: permutation test of the diagonal mean.}
  \label{fig:similarity}
\end{figure*}

\paragraph{Measured alignment.}
We place each training sample on the update map of \Cref{fig:update-map}, using the judge midway through training with domain reweighting (\Cref{fig:update-samples}). The alignment of a sample is $\langle \nabla_\phi m_i, -P g_{\text{meta}}\rangle$ of \Cref{eq:factor} with $P = I$, computed from its GRPO gradient over 8 judgments (which equals $-\psi_i\nabla_\phi m_i$) and the mean gradient of the meta set. Easy problems move their weight the least: 40\% carry no signal, and the weight rises for 45\%, against 62\% of medium and 60\% of hard problems. Weighted by the learning signal, the mean alignment of hard problems is 1.26 times that of easy problems, and the mean of $\psi_i$ times the alignment is 1.48 times; the two ratios stay within 1.20--1.29 and 1.41--1.51 when the meta set is restricted to its easy, medium or hard problems. Medium and hard problems align alike.

\begin{figure*}[p]
\begin{promptframe}[label={lst:rrm-prompts}]{Judge prompt for reasoning-based selection}
\begin{minipage}[t]{0.485\linewidth}
\begin{lstlisting}[style=prompt]
=== PROGRAMMING PROBLEM ===

Title: {question_title}
Platform: {platform}
Difficulty: {difficulty}

Problem Statement:
{question_content}

Starter Code:
```python
{starter_code}
```

=== PUBLIC TEST CASES ===

The following are example test cases that illustrate the problem.
IMPORTANT: These are NOT the complete test suite. Hidden test cases will include edge cases, boundary conditions, and special inputs not represented below.

{test_cases}

=== CANDIDATE SOLUTION ===

--- Solution 1 ---

```python
{solution_code}
```

--- Solution 2 ---

```python
{solution_code}
```

=== SELECTION TASK ===

You are given {num_solutions} candidate solutions. Select the BEST solution that is most likely to pass ALL test cases (public examples and hidden test cases).

The public test cases you see are merely examples. The actual test suite contains hidden test cases specifically designed to catch edge cases, boundary conditions, and algorithmic errors that may not appear in examples.

Be skeptical and rigorous. Select the solution you are most confident will handle all possible valid inputs.

\end{lstlisting}
\end{minipage}\hfill
\begin{minipage}[t]{0.485\linewidth}
\begin{lstlisting}[style=prompt]
EVALUATION PROTOCOL (apply to each solution):

STEP 1: SYNTAX & COMPLETENESS VERIFICATION
- Is the code syntactically valid?
- Is the code complete?
- Are all variables defined before use? Are all imports present?
- Are function signatures complete with all parameters and return statements?

STEP 2: ALGORITHM CORRECTNESS ANALYSIS
- Is the chosen algorithm correct for this problem type?
- Walk through the logic line-by-line: Are operators correct? Are loop bounds correct? Are conditionals correct?
- Check for common errors: off-by-one errors, integer overflow, wrong comparison operators (< vs <=), incorrect loop termination
- Does the algorithm match the problem requirements exactly?

STEP 3: EDGE CASE ENUMERATION & TESTING
The public test cases show only "happy path" examples. Mentally test against these comprehensive edge case categories which apply to the problem:

- SIZE EDGE CASES: Empty input, Minimum size, Maximum size, Boundary sizes
- VALUE EDGE CASES: Minimum values, Maximum values, Zero, Negative numbers, Duplicates
- STRUCTURAL EDGE CASES: Sorted arrays, Reverse-sorted arrays, All elements identical, Alternating patterns, Single unique element among duplicates
- PROBLEM-SPECIFIC EDGE CASES: For graphs: disconnected components, self-loops, cycles; For strings: single character, all same character, alternating characters; For trees: single node, linear tree (linked list), complete tree; For intervals: overlapping, touching, disjoint, nested

STEP 4: INPUT/OUTPUT FORMAT VERIFICATION
- Input parsing: Does code read exactly the specified format?
- Output format: Exact spacing? Exact newlines? Correct precision for floats? Correct ordering of multiple outputs?

STEP 5: RUNTIME SAFETY & PERFORMANCE
Check for errors that appear during execution: Division by zero, Index out of bounds, Infinite loops, Time complexity, Space complexity, Stack overflow

OUTPUT FORMAT:
Provide step-by-step analysis following the 5-step protocol above for each solution, then end your response with exactly:
<selection>INDEX</selection>

where INDEX is 1 to {num_solutions}, indicating the solution most likely to pass ALL test cases.
\end{lstlisting}
\end{minipage}
\end{promptframe}
\end{figure*}

\begin{figure*}[!t]
  \centering
  \includegraphics[width=\textwidth]{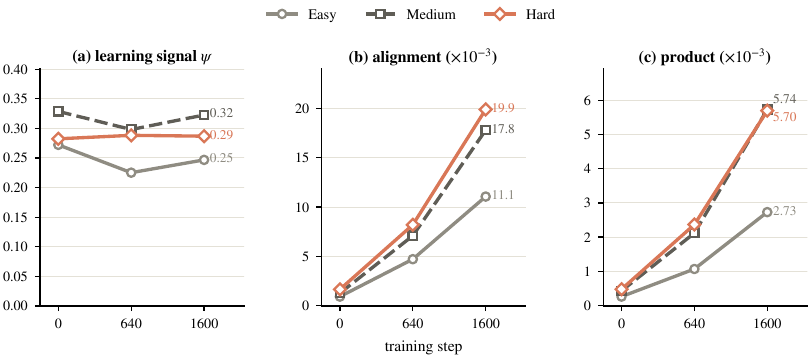}
  \caption{(a) Learning signal $\psi$, (b) alignment and (c) their product by difficulty over training, for a run of domain reweighting whose meta set holds only hard problems ($P = I$). The alignment of a difficulty is the $\psi$-weighted mean alignment of its samples, and the product is the mean of $\psi_i$ times the alignment.}
  \label{fig:factor-dynamics}
\end{figure*}

\paragraph{Task and trajectory similarity.}
The alignment also separates training samples by their task and by their trajectories (\Cref{fig:similarity}). We group the training samples and the meta samples by platform, for the task, and by the policy model that produced the candidates, for the trajectory; for this grouping we also judge DeepSeek V3.2 Speciale candidates of the training problems, so that both sides share a source. Every training gradient aligns with the mean gradient of the whole meta set, so we remove this common direction from each training gradient and from the mean meta gradient of each group, and take the cosine of what remains. We then double-center each matrix: a cell gives how much a training group aligns with one meta group beyond its alignment with every meta group. The diagonal is positive in both panels, so training samples align more with meta samples of their own platform and source. The excess is clear for the candidate source and weaker for the platform. For the judge midway through training with domain reweighting, the same-source excess is $+0.025$ (95\% interval over problems $[+0.010, +0.040]$, permutation $p < 0.001$), largest for DeepSeek V3.2 Speciale, the stronger policy model. The same-platform excess is $+0.019$ (permutation $p = 0.006$; the interval over problems, $[-0.004, +0.041]$, includes zero); it holds on easy problems ($+0.022$ $[+0.007, +0.040]$).

\paragraph{Change over training.}
Both factors change as the judge trains (\Cref{fig:factor-dynamics}). We repeat the measurement at steps 0, 640 and 1600 of a run of domain reweighting whose meta set holds only hard problems. The learning signal stays between 0.22 and 0.33, while the alignment of every difficulty grows about tenfold. Hard problems have the highest alignment at every step, 1.7 to 1.8 times that of easy problems, and easy problems have the lowest product at every step.

\section{Prompt Templates}
\label{appendix:prompts}

The judge prompt, shown in \Cref{lst:rrm-prompts}, instructs the model to follow a structured five-step evaluation protocol before selecting the best candidate solution.
The five steps are: (1)~\emph{Syntax \& Completeness Verification}, which checks that code is syntactically valid, complete, and has all necessary imports and function signatures; (2)~\emph{Algorithm Correctness Analysis}, which walks through the logic line by line, looking for off-by-one errors, wrong operators, and other common mistakes; (3)~\emph{Edge Case Enumeration \& Testing}, which mentally tests the code against comprehensive categories of boundary inputs beyond the public examples; (4)~\emph{Input/Output Format Verification}, which confirms the code reads and writes data in the exact format required; and (5)~\emph{Runtime Safety \& Performance}, which checks for division by zero, index-out-of-bounds, infinite loops, and excessive time or space complexity.
They cover well-formedness, algorithmic soundness, robustness, format compliance, and efficiency; the judge writes its analysis before selecting (\Cref{sec:preliminary}).

\end{document}